\documentclass[journal]{IEEEtran}
\ifCLASSINFOpdf
\else
\fi
\usepackage{xcolor}
\usepackage{graphicx} 

\usepackage{algorithm}
\usepackage{algorithmic}

\usepackage{tikz}
\usepackage{amsmath} 
\usepackage{bbding}
\usepackage{makecell}

\usepackage{xpatch}
\usepackage{colortbl}  
\usepackage{array} 
\usepackage{amssymb}
\usepackage{booktabs, multirow, tabularx}

\usepackage[colorlinks, citecolor=blue, linkcolor=red, urlcolor=magenta]{hyperref}

\definecolor{mygray}{gray}{.9}

\definecolor{clblue}{RGB}{222, 246, 246}
\newcommand{\cc}[1]{\cellcolor{clblue!70}{#1}}

\definecolor{cgray}{RGB}{191, 191, 191}

\definecolor{clblue}{RGB}{209, 246, 246}

\definecolor{clorange}{RGB}{255, 136, 16}

\definecolor{tabletitle}{gray}{.8}
\definecolor{ours}{gray}{.95}
\definecolor{ggray}{RGB}{127,127,127}

\definecolor{reda}{RGB}{202,0,0}

\definecolor{redb}{RGB}{217,148,143}

\definecolor{myyellow}{RGB}{190,144,0}
\definecolor{mygreen}{RGB}{0,136,51}
\definecolor{myblue}{RGB}{0,102,204}
\newcolumntype{B}{!{\vrule width 1pt}}
\usepackage{pifont}

\begin{document}

%
\title{When Color-Space Decoupling Meets Diffusion for Adverse-Weather Image Restoration}
%
%
%

\author{Wenxuan Fang, Jili Fan, Chao Wang, Xiantao Hu, Jiangwei Weng, Ying Tai, Jian Yang\textsuperscript{\rm *} and Jun Li\textsuperscript{\rm *}

\thanks{}
\thanks{ }
\thanks{}

\thanks{This work was supported by the National Natural Science Foundation of China under Grant Nos. U24A20330, 62361166670.}
\thanks{Wenxuan Fang, Chao Wang, Xiantao Hu, Jiangwei Weng, Jian Yang and Jun Li are with the School of Computer Science and Engineering, Nanjing University of Science and Technology, Nanjing, 210000, China (email: \{wenxuan\_fang, cw0601, yuzheng, xiantaohu, wengjiangwei, csjianyang, junli\}@njust.edu.cn)}
\thanks{Jili Fan is with the School of Electrical Engineering, Southeast University, Nanjing, 214135, China (e-mail: 230258432@seu.edu.cn).}
\thanks{Ying Tai is with School of Intelligence Science and Technology, Nanjing University, Suzhou, 215009, China. (e-mail: yingtai@nju.edu.cn).}

\thanks{Corresponding authors: Jian Yang and Jun Li.}

\thanks{* indicates Corresponding authors.}

}

\maketitle

\begin{abstract}
Adverse Weather Image Restoration (AWIR) is a highly challenging task due to the unpredictable and dynamic nature of weather-related degradations. 
Traditional task-specific methods often fail to generalize to unseen or complex degradation types, while recent prompt-learning approaches depend heavily on the degradation estimation capabilities of vision-language models, resulting in inconsistent restorations. 
In this paper, we propose \textbf{LCDiff}, a novel framework comprising two key components: \textit{Lumina-Chroma Decomposition Network} (LCDN) and \textit{Lumina-Guided Diffusion Model} (LGDM). LCDN processes degraded images in the YCbCr color space, separately handling degradation-related luminance and degradation-invariant chrominance components. This decomposition effectively mitigates weather-induced degradation while preserving color fidelity. 
To further enhance restoration quality, LGDM leverages degradation-related luminance information as a guiding condition, eliminating the need for explicit degradation prompts. Additionally, LGDM incorporates a \textit{Dynamic Time Step Loss} to optimize the denoising network, ensuring a balanced recovery of both low- and high-frequency features in the image. Finally, we present DriveWeather, a comprehensive all-weather driving dataset designed to enable robust evaluation. Extensive experiments demonstrate that our approach surpasses state-of-the-art methods, setting a new benchmark in AWIR. The dataset and code are available at: \url{https://github.com/fiwy0527/LCDiff}.
\end{abstract}


%
\IEEEpeerreviewmaketitle

\section{Introduction}

\IEEEPARstart{A}{dverse} Weather Image Restoration (AWIR) focuses on recovering the clarity and fidelity of images degraded by various weather conditions, aiming to produce results that are both visually coherent and semantically consistent. Weather phenomena such as rain, haze, and snow often diminish image contrast, obscure fine details, and introduce unwanted artifacts, all of which can significantly impair the performance of vision-based tasks, including object tracking \cite{hu2024exploiting, hu2023transformer}, object detection \cite{wang2023category, wang2025msod} and image segmentation \cite{segment, chen2024transformer}.

\begin{figure}[!t]
    \centering 
    \includegraphics[width=0.97\linewidth]{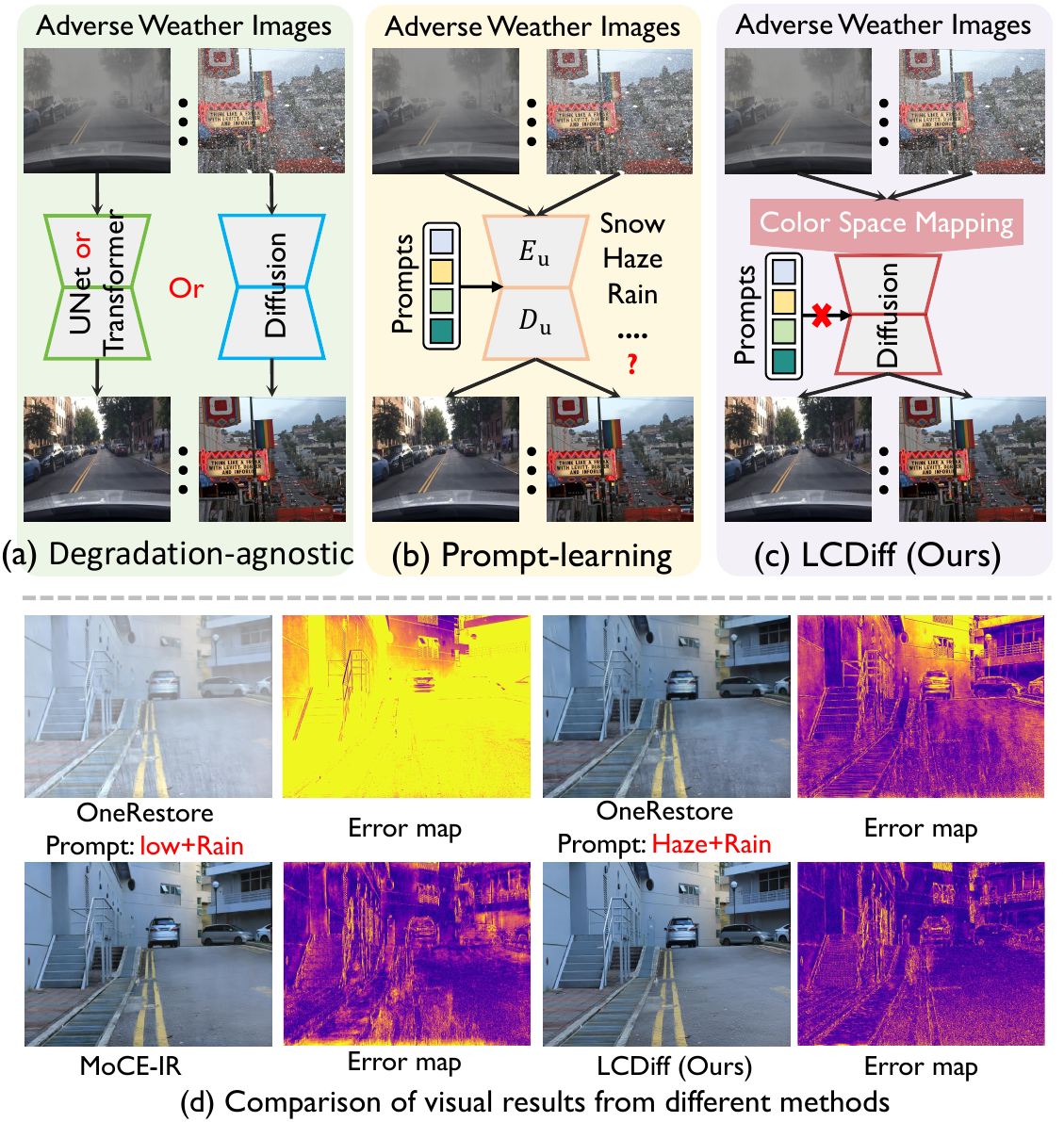}
   
    \caption{Representative frameworks of adverse weather image restoration. (a) Degradation-agnostic approaches restore diverse weather with a unified model but typically operate in RGB where intensity and chroma are entangled. (b) Prompt-learning methods utilize text prompts or learnable queries for task adaptability. (c) Our LCDiff handles multi-weather in YCbCr space without degradation prompts. (d) Visual comparison between OneRestore \cite{guo2025onerestore}, MoCE-IR \cite{MoCEIR} and our approach under mixed conditions of rainfog, and glare.} 
    \label{figure:overview}
  
\end{figure}

\begin{figure*}[!t]
    \centering 
    \includegraphics[width=0.97\linewidth]{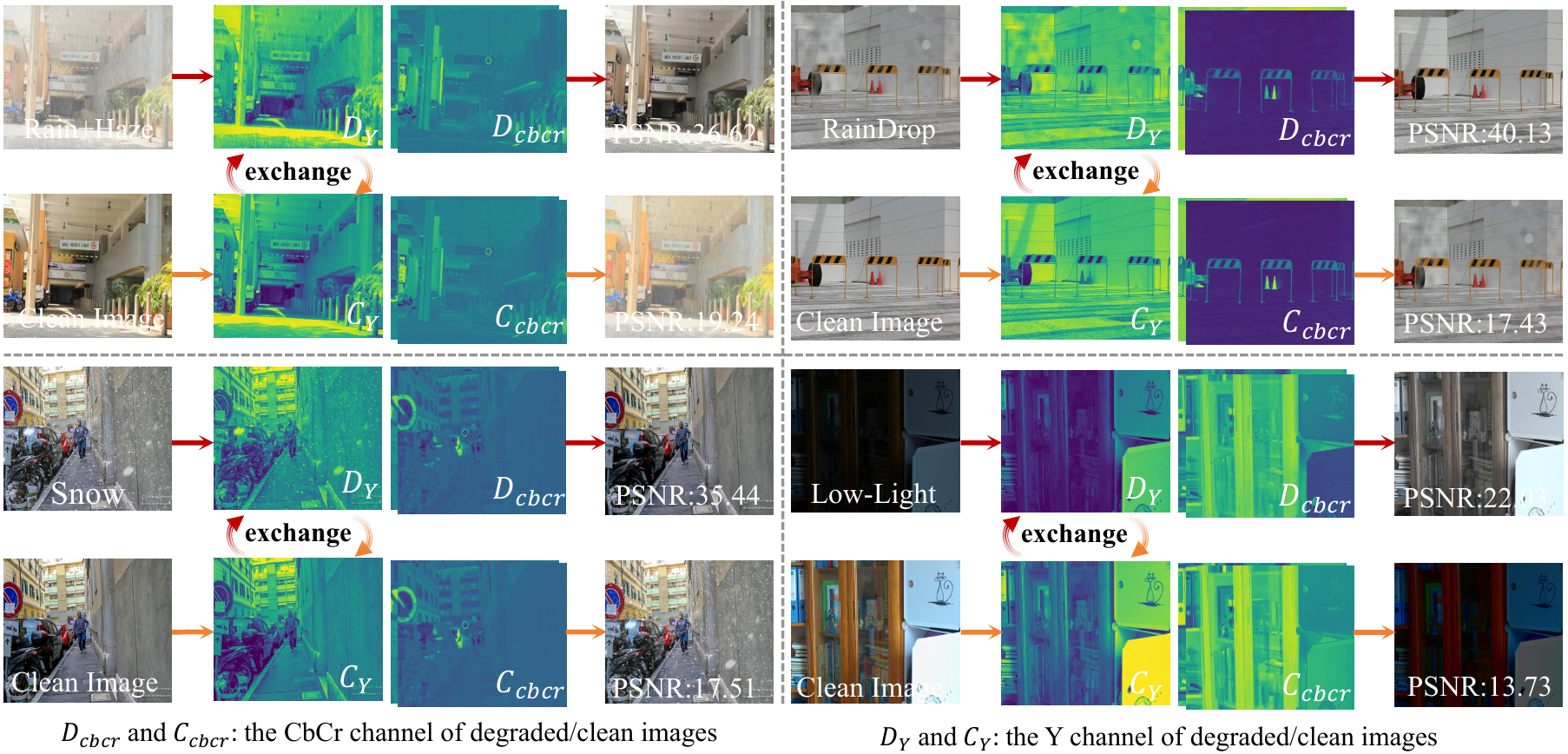}
    \caption{In the YCbCr color space, weather-induced degradations predominantly affect the luminance (Y) channel, with only minor blurring occurring in the Chroma (CbCr) channels. By swapping the Y channel of a degraded image with that of a clean image, we can obtain a clear and faithful restoration, even in images with coexisting rain and fog.}
    \label{figure:motivation}
   
\end{figure*}

Existing AWIR methods \cite{chen2025teaching, potlapalli2024promptir, MoCEIR, lin2025diffbir, ozdenizci2023restoring, cui2025adair, valanarasu2022transweather, yu2024multi} are broadly divided into \textit{degradation-agnostic} models \cite{cui2025adair,ozdenizci2023restoring, valanarasu2022transweather, xu2025unifying} and \textit{prompt-learning} approaches \cite{chen2025teaching, MoCEIR, potlapalli2024promptir}. Degradation-agnostic models train a single network that maps diverse degradations into a shared representation space to expose cross-weather commonality (Figure~\ref{figure:overview}(a)). However, as degradation diversity and volume grow, maintaining both compactness and separability in this space becomes challenging, risking representation collapse and degraded performance on long-tail cases. Prompt-learning methods typically leverage pre-trained vision--language (PVL) models \cite{lai2024lisa} to generate weather-specific prompts and thereby handle unseen degradations \cite{chen2025teaching, potlapalli2024promptir} (Figure~\ref{figure:overview}(b)). While effective in controlled settings \cite{guo2025onerestore, luo2024controlling, MoCEIR}, they are brittle to prompt estimation: inaccurate or inconsistent prompts yield uncontrollable restoration quality. As illustrated in Figure~\ref{figure:overview}(d), OneRestore \cite{potlapalli2024promptir} produces inconsistent prompts under overlapping rain and fog, resulting in obvious recovery errors, which is disastrous for practical applications.


To develop a robust all-in-one model capable of effectively addressing the AWIR task, we begin with a strategic experiment by converting RGB images—degraded by haze, rain, low-light, and snow—into the YCbCr color space in Figure~\ref{figure:motivation}. By swapping the luminance (Y) channels between degraded and clean images, we observed that degraded images with the clean Y channel exhibit higher PSNR values, whereas clean images with the degraded Y channel experience a significant drop in PSNR.
This suggests that weather-induced degradations predominantly affect the Y channel, while the chrominance (CbCr) channels remain largely intact, showing only minor blurring. 
This key insight drives our core approach to decompose degraded images into a degradation-concentrated Y map and degradation-invariant CbCr maps, enabling more effective recovery of the Y component. Simultaneously, leveraging the Y map as a conditioning signal in the diffusion process eliminates the need for precise weather prompts, enhancing the model's ability to handle diverse and complex real-world degradations.


Based on the aforementioned analysis, we develop a unified restoration diffusion model, \textbf{LCDiff}, built around two complementary components. First, the \textbf{Lumina–Chroma Decomposition Network (LCDN)} is proposed  operates in YCbCr to disentangle degradation-concentrated luminance from degradation-invariant chroma, removing weather artifacts in Y while preserving the structural coherence of Cb/Cr. This projection shrinks the restoration search space and mitigates RGB entanglement, yielding a coarse yet semantically faithful input to diffusion. Second, the \textbf{Lumina-Guided Diffusion Model (LGDM)} conditions denoising on the restored Y, stabilizing reverse trajectories and explicitly steering the model toward high-frequency recovery that vanilla diffusion under-recovers \cite{san2023discrete,han2024asyncdsb}. To counter the frequency–SNR imbalance inherent to DDPMs, LGDM employs a \emph{dynamic time-step loss} that emphasizes low-frequency consistency early and progressively shifts weight to high-frequency details, improving both texture fidelity and convergence. Unlike prompt-driven pipelines, LCDiff removes dependence on brittle PVL prompts, remains robust under overlapping degradations, and is computationally efficient by avoiding full diffusion on chroma. Finally, we introduce \textbf{DriveWeather}, a comprehensive driving benchmark spanning seven adverse weather conditions (e.g., haze, rain–fog, glare), enabling rigorous, reproducible evaluation of unified AWIR systems. In summary, our contributions are as follows:

\begin{figure*}[!t]
    \centering 
    \includegraphics[width=0.97\linewidth]{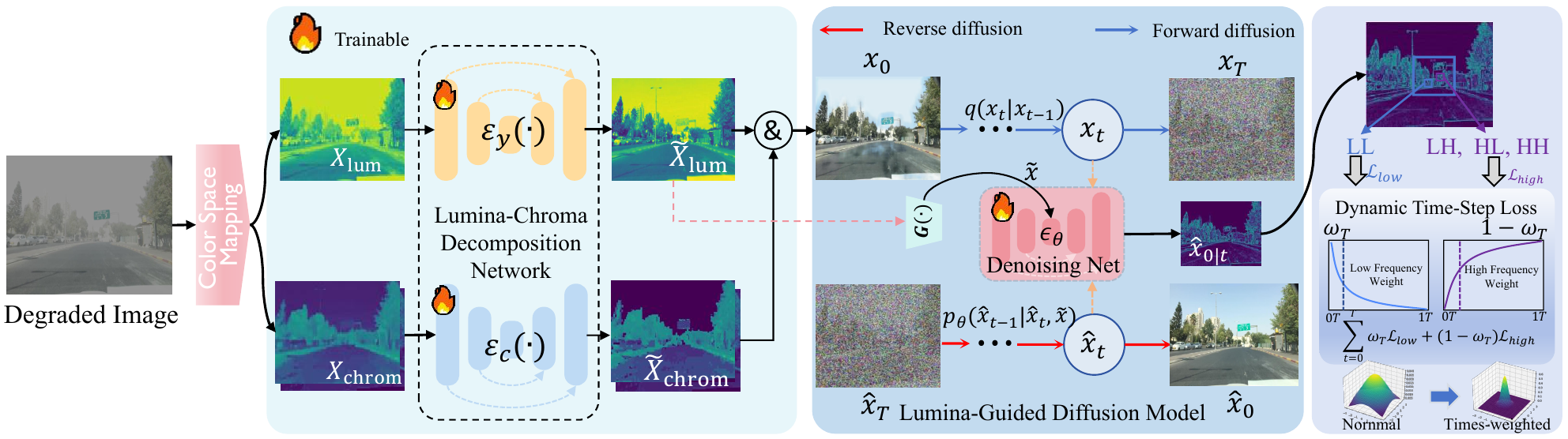}
  
    \caption{Pipeline of our LCDiff including LCDN and LGDM. LCDN is responsible for removing weather degradation in the YCbCr color space, while LGDM enhances the restoration process by adaptively refining both low and high-frequency details.} 
    \label{figure:framework}

\end{figure*}

\begin{itemize}
    \item We propose a Lumina-Chroma Decomposition Network (LCDN) that effectively separates degradation-related luminance from degradation-invariant chrominance, enabling unified restoration of weather-degraded images.  
    \item We propose a Lumina-Guided Diffusion Model (LGDM) that utilizes the decomposed luminance map as a conditional input, eliminating the need for degradation prompts. Additionally, we design a dynamic time step loss to enhance the quality of the restored images.
    \item By integrating LCDN with LGDM, we present a robust LCDiff for weather-induced image restoration. To the best of our knowledge, this is the first work to incorporate the YCbCr color space within a diffusion model for unified weather degradation removal.
    \item We create DriveWeather, a comprehensive driving scenes dataset for all-weather image restoration, and demonstrate through extensive experiments that our method outperforms related state-of-the-art techniques.
\end{itemize}

\section{Related Work}
\subsection{Adverse Weather Image Restoration.}
Early works in AWIR primarily focused on specific weather degradations, such as fog \cite{zheng2023curricular, zhang2023sdbad}, rain \cite{chen2024bidirectional}, and snow \cite{liu2018desnownet}. In response, degradation-aware methods have emerged, including neural architecture search \cite{li2020all}, knowledge distillation \cite{chen2022learning}, and learnable queries \cite{valanarasu2022transweather}. For example, a pioneering work in this direction, All-In-One \cite{li2020all}, employed multiple task-specific encoders and neural architecture search to address a range of weather degradations. Following this, Transweather \cite{valanarasu2022transweather} introduced learnable weather category queries that effectively removed weather-induced degradations. Later, WGWS-Net \cite{zhu2023learning} focused on both general and weather-specific features. Other notable contributions include the application of clustering \cite{cao2024grids} and diffusion models \cite{lin2025diffbir, chen2025teaching} in AWIR. In parallel, recent works have explored the potential of prompt learning \cite{chen2025teaching, dppd} in AWIR, including visual \cite{guo2025onerestore}, textual \cite{luo2024controlling}, and learnable queries \cite{potlapalli2024promptir}. For instance, OneRestore \cite{guo2025onerestore} utilized both visual and textual prompts to enable adaptive restoration under complex weather conditions, while MoCE-IR \cite{MoCEIR} introduce a
complexity-aware allocation mechanism that preferentially
directs tasks to lower-complexity experts for handling multiple weather conditions. In contrast, our LCDiff approach decouples degraded visual content by decomposing images into luminance and chrominance components, enabling independent recovery and unified weather degradation removal—without relying on linguistic dependencies.

\subsection{Color Space in Image Restoration.}
Beyond the standard RGB color model, several studies have explored the use of alternative color spaces, such as HSV \cite{lyu2024mcpnet}, YCbCr \cite{guo2023low, fang2024guided}, and HVI \cite{yan2025hvi}, for specific image restoration tasks like low-light enhancement \cite{she2024mpc} and underwater image enhancement \cite{liu2023multi}. For example, Bread \cite{guo2023low} leveraged the YCbCr color space for noise reduction and color correction in low-light image restoration. SGDN \cite{fang2024guided} integrated YCbCr features to guide RGB features, improving real-world image dehazing. MSDC-Net \cite{liu2023multi} adaptively selected features from both RGB and HSV color spaces to enhance underwater image restoration quality. While these methods demonstrate excellent performance within their specific tasks, their specialized architectures pose challenges in adapting them to unified AWIR. In contrast to prior approaches, we observe that consistent patterns in the YCbCr color space effectively enable the unified removal of weather-induced degradations.

\section{Methodology}
\label{methods}

In this section, we propose LCDiff, a novel framework designed to address complex weather-induced image degradations without relying on explicit degradation prompts, illustrated in Figure~\ref{figure:framework}. The framework consists of two core components: Lumina-Chroma Decomposition Network and Lumina-Guided Diffusion Model.


\subsection{Lumina-Chroma Decomposition Network}
Given an unknown degraded RGB image \( X_{d} \in \mathbb{R}^{H \times W \times 3} \), a color space conversion to obtain its YCbCr space. This transformation separates the degraded image into two distinct components: a degradation-related luminance map \( X_{\text{lum}} \in \mathbb{R}^{H \times W \times 1} \) and a degradation-invariant chrominance map \( X_{\text{chrom}} \in \mathbb{R}^{H \times W \times 2} \). Our LCDN comprises two independent subnetworks: the Luminance Restoration Module \(\varepsilon_{y}(\cdot)\), the Frequency Chrominance Restoration Module \(\varepsilon_{c}(\cdot)\) in Figure~\ref{figure:framework}. Specifically, \(\varepsilon_{y}(\cdot)\) processes the degradation-related luminance map $X_{\text{lum}}$, ensuring semantic consistency and preserving image details, while \(\varepsilon_{c}(\cdot)\) processes the degradation-invariant chrominance map $X_{\text{chrom}}$, maintaining color authenticity and consistency. They are formalized as: 
\begin{equation}
    \widetilde{X}_{\text{lum}} = \varepsilon_{y}(X_{\text{lum}}; \theta_{Y}), \ \widetilde{X}_{\text{chrom}} = \varepsilon_{c}(X_{\text{chrom}}; \theta_{C}),
    \label{eq:lcdn}
\end{equation}
where $\widetilde{X}_{\text{lum}}$ and $\widetilde{X}_{\text{chrom}}$ are the restored luminance and chrominance, respectively, and $\theta_{Y}$ and $\theta_{C}$ denote the subnetwork parameters. These restored maps are then concatenated, generating the initial input \( x_0 \) for the subsequent diffusion model. 
At this stage, LCDN aims to eliminate degradation information from the low-quality image without imposing additional generative complexity on the subsequent diffusion process.

\begin{figure}[!t]
    \centering 
    \includegraphics[width=1\linewidth]{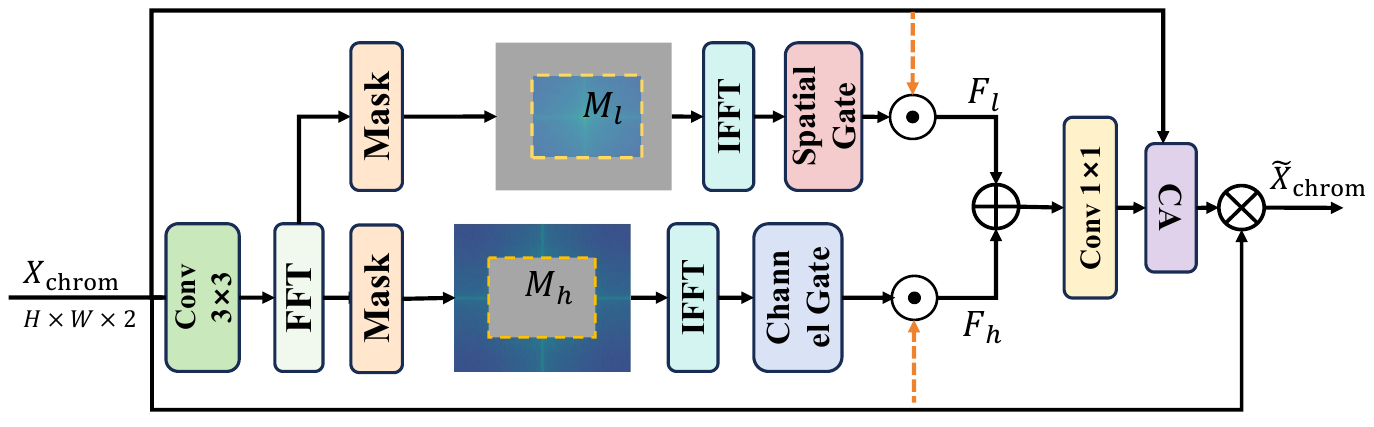}
    \caption{The framework of Frequency Chrominance Restoration Module.} 
    \label{figure:fcrm}
\end{figure}

\textbf{Luminance Restoration Module (LRM).} 
It specifically handles the luminance information in the degraded image. In particular, the luminance is the most affected by degradation. Leveraging NAFNet's \cite{chen2022simple} outstanding performance in image restoration, we employ its pixel and channel attention mechanisms to effectively capture long-range dependencies, leading to better recovery of image details, especially structural information. It is worth noting that established pre-trained models and frameworks, such as NAFNet \cite{chen2022simple} and FSNet \cite{cui2023image}, can be used in luminance recovery, achieving excellent restoration results without the need for unique designs.

\textbf{Frequency Chrominance Restoration Module (FCRM).} 
In natural images, chrominance components, often represented in YCbCr color spaces primarily encode color information. Conventional spatial-domain restoration methods often struggle to balance global chromatic consistency and local detail preservation, leading to artifacts such as color bleeding or texture oversmoothing. To address this issue, we design a Frequency Chrominance Restoration Module, which decomposes chrominance signals into low and high-frequency components for targeted refinement during the diffusion process.

As shown in Figure \ref{figure:fcrm}, we perform frequency decomposition via a 2D Fourier Transform (FFT) \cite{nussbaumer1982fast}: $X_{f} = \mathcal{F}(X_{\text{chorm}})$, allowing explicit separation of low and high-frequency chrominance components:
\begin{equation}
     \quad X_{l} = \mathcal{F}^{-1}(M_{l} \odot X_{f}), \quad X_{h} = \mathcal{F}^{-1}(M_{h} \odot X_{f}),
\end{equation}
where \(\mathcal{F}\) and \(\mathcal{F}^{-1}\) represent the forward and inverse Fourier transforms, respectively, and \(M_{l}\), \(M_{h}\) denote low and high-frequency masks, ensuring non-overlapping decomposition (i.e., \(M_{l} + M_{h} = 1\)). For low-frequency chrominance restoration, we employ a spatial attention mechanism to enhance global color consistency while suppressing noise. This is achieved via a gating function:
\begin{equation}
    S_{l} = \sigma(W_{s} \ast \text{cat}(\max(X_{l}), \mu(X_{l}))),
\end{equation}
where \(W_{s}\) is a convolutional filter, \(\max(X_{l})\) and \(\mu(X_{l})\) represent the channel-wise max and mean operations, respectively, and \(\sigma(\cdot)\) is the sigmoid activation. 

For high-frequency chrominance, we introduce a channel attention mechanism to amplify fine details and prevent over-smoothing. The attention score is computed as:
\begin{equation}
    S_{h} = \sigma(W_{c2} \ast \text{ReLU}(W_{c1} \ast (\text{AvgPool}(X_{h}) + \text{MaxPool}(X_{h})))),
\end{equation}
where \(W_{c1}\) and \(W_{c2}\) are learnable convolutional layers, and both average and max pooling functions help extract global and local contextual features. By modulating \(X_{h}\) with \(S_{h}\), the model selectively enhances significant high-frequency chrominance features while mitigating unwanted noise. We then splice the high and low frequency outputs together:
\begin{equation}
    F_{l}=S_{l} \odot X^{chorm}, \quad F_{h}=S_{h} \odot X^{chorm},
\end{equation}
\begin{equation}
    F_{chorm} = F_{h} +  F_{l}.
\end{equation}

Since chrominance restoration requires multi-scale contextual understanding, we employ a Channel Cross-Attention (CA) mechanism to integrate global and local features:
\begin{equation}
    Q = W_q \ast F_{chorm}, \quad K, V = \text{split}(W_k \ast X_{\text{chorm}})
\end{equation}
where \( W_q \) and \( W_k \) are convolutional projections. The attention scores are computed as:
\begin{align}
z = W_o \ast (\alpha V), \ \ \
\alpha = \text{softmax} \left( \frac{Q K^T}{\sqrt{d}} \right),    
\end{align}
where \( d \) is the feature dimension. \( W_o \) is a projection layer. This operation enhances the chrominance feature representation by dynamically attending to spatially relevant chrominance information across channels. 
Finally, we modulate the chroma multiplicatively and obtain the final chrom $\widetilde{X}_{\text{chrom}} = z\otimes X_{\text{chrom}}$.

\begin{figure}[t]
    \centering 
    \includegraphics[width=0.97\linewidth]{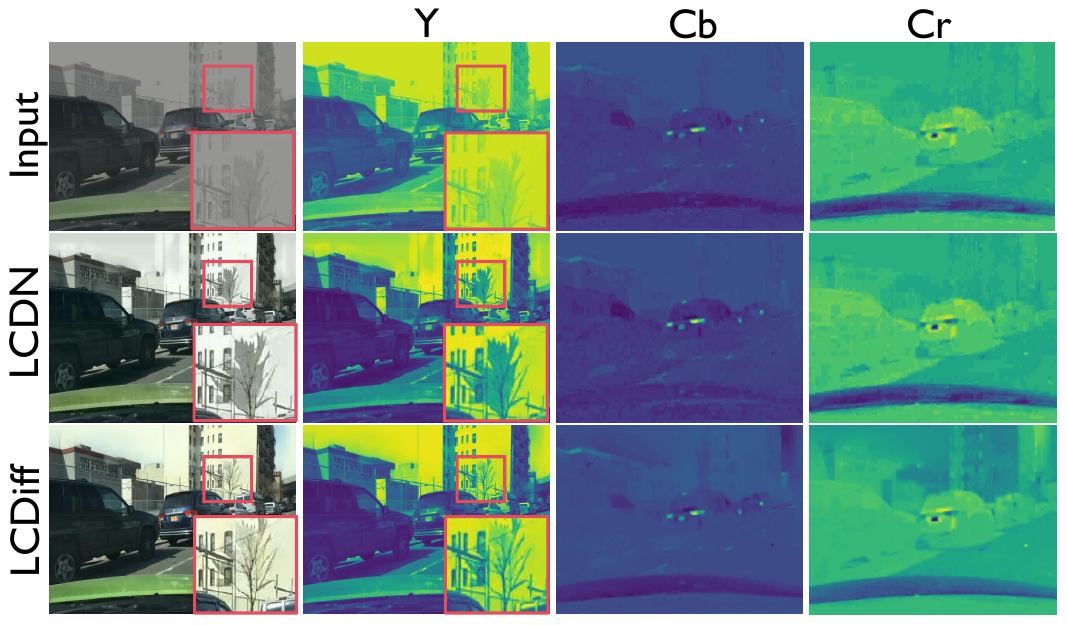}
    \caption{Comparison between LCDN and LCDiff. LCDN effectively removed degradation but caused some texture distortion. LCDiff more accurately restored the textures. } 
    \label{figure:lcdn}

\end{figure}

\textbf{Training loss.} The optimization objective is defined as: $\mathcal{L}_{res} =  \eta \mathcal{L}_{\ell_1} + \theta \mathcal{L}_\text{ssim} + \lambda \mathcal{L}_\text{fft}$, where \( \mathcal{L}_{\ell_1} \) is the $\ell_1$ loss, \( \mathcal{L}_\text{ssim} \) the structural similarity loss (enhancing perceptual quality), \( \mathcal{L}_\text{fft} \) is the frequency-domain loss. The parameters \( \eta \), \( \theta \), and \( \lambda \) are set to 1.0, 0.5, and 0.1, respectively.

\subsection{Lumina-Guided Diffusion Model}
\label{sec:LGDM}
After implementing the LCDN in the previous subsection, the degradation of the luminance map is effectively mitigated. However, certain challenging scenarios, such as excessive blurring in the luminance map, can still lead to artifacts in the restored image (see Figure~\ref{figure:lcdn}). To overcome this limitation, we propose the Lumina-Guided Diffusion Model (LGDM), which integrates the generative capabilities of diffusion models with the guiding influence of luminance information, thereby enhancing the recovery of image content.

\textbf{Forward Diffusion.} Starting from a coarsely restored image \cite{lightendiffusion, chung2022come} brings the initial distribution closer to the clean target distribution, enabling faster convergence and requiring fewer sampling steps. In the forward diffusion process, noise is incrementally added to the initial image \( x_0 = cat(\widetilde{X}_{lum}, \widetilde{X}_{chrom})\), as defined in Eq. \eqref{eq:lcdn}, over \( T \) time steps. This process is modeled as a Markov chain, where noise is progressively injected at each step:
\begin{equation}
    q(x_t | x_{t-1}) = \mathcal{N}(x_t; \sqrt{1-\beta_t} x_{t-1}, \beta_t \mathbf{I}),
    \label{eq:addnoise}
\end{equation}
where \( x_t \) denote  the noisy image at time step $t$, and \( \beta_t \) is the variance schedule controlling the noise level.

\begin{figure*}[!t]
    \centering 
    \includegraphics[width=0.95\linewidth]{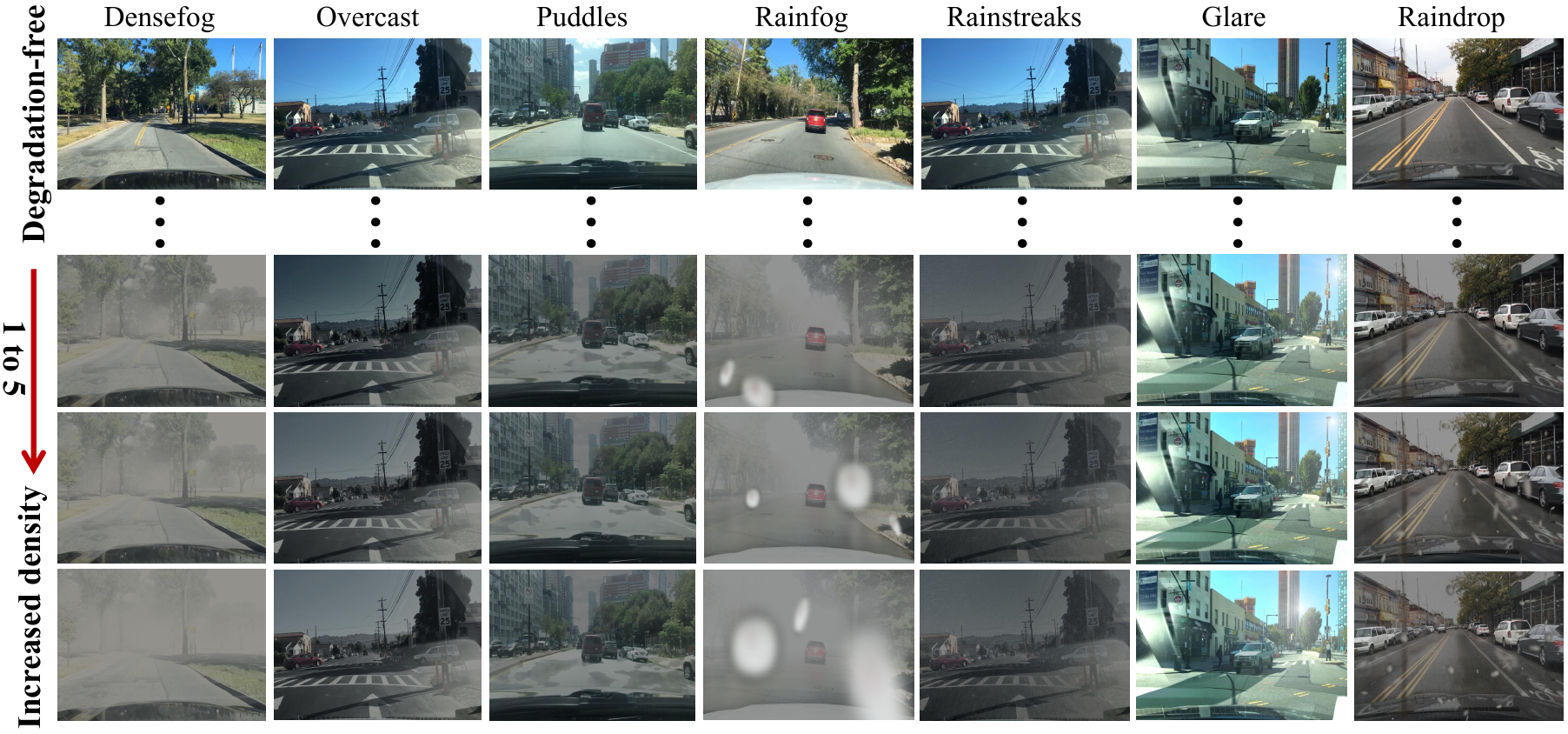}
    \caption{Example images of our DriveWeather dataset. It includes 29,750 pairs of clean and degraded images. Each column shows a specific weather scenario.} 
    \label{figure:driverweather_dataset}
\end{figure*}




\textbf{Reverse Denoising.} In the reverse denoising process, our goal is to recover the clean image \( \hat{x}_0 \), starting from the noisy image \( \hat{x}_T \). However noise and degradation frequently introduce instability into the generation process. To address this, we utilize conditional diffusion \cite{chung2022come}, injecting the restored luminance condition \( \textcolor{red}{\widetilde{x}} = G(\widetilde{X}_{lum}) \) to refine the denoising process. This strategy enables the network to initiate from a luminance component that has already undergone partial restoration, thereby guiding a more stable generation process. It capitalizes on inherent, deterministic prior knowledge to mitigate unnecessary noise and unnatural artifacts. This process is formally characterized by conditional probability:
\begin{equation}
    p_{\theta}(\hat{x}_{t-1}| \hat{x}_t, \textcolor{red}{\widetilde{x}}) = \mathcal{N}(\hat{x}_{t-1}; \mu_\theta(\hat{x}_t, t, \textcolor{red}{\widetilde{x}}), \sigma_t^2 \mathbf{I}),
\end{equation}
where \( \sigma_t^2 \) is the variance, and \( \mu_\theta(\hat{x}_t, t, \widetilde{x}) \) is the mean value. 

\textbf{Training Process.} Here, the overall loss to train the deniosing network in LGDM is defined as follows: $\mathcal{L}_{\text{all}}=\mathcal{L}_{\text{denoise}}+\mathcal{L}_{\text{dts}}$, where \(\mathcal{L}_{\text{denoise}}\) a DDPM loss and \(\mathcal{L}_{\text{dts}}\) is a dynamic time step loss.  

\textbf{\textit{DDPM loss.}} At each time step \( t \), the model predicts the noise \( \varepsilon_{\theta}(\hat{x}_t, t, \widetilde{x}) \), and the objective is to minimize the difference between the predicted and true noise:
\begin{equation}
    \mathcal{L}_\text{denoise} = \mathbb{E}_{\hat{x}_0, t}  \left\| \varepsilon_{\theta}(\hat{x}_t, t, \widetilde{x}) - \varepsilon_t \right\|_2^{2},
\end{equation}
where \( \varepsilon_t \) is the true noise term at time step \( t \).

\textbf{\textit{Dynamic Time Step (DTS) Loss.}} Previous studies \cite{han2024asyncdsb, san2023discrete} have shown that diffusion models often exhibit a bias toward restoring low-frequency regions due to the influence of synchronized noise scheduling. This bias delays the recovery of high-frequency details, ultimately degrading the fine-grained features of the restored images. Here, we introduce a DTS loss, which adaptively balances the restoration of low-frequency and high-frequency features by dynamically adjusting its focus based on the current time step. The DTS loss is defined as:
\begin{equation}
    \mathcal{L}_{\text{dts}} = \omega_t \mathcal{L}_{\text{low}}(\hat{x}_{0|t}^{l}, x_{0}^{l}) + (1 - \omega_t) \mathcal{L}_{\text{high}}(\hat{x}_{0|t}^{h}, x_{0}^{h}),
\end{equation}
where \( \omega_{t} = e^{-k \times \frac{t}{T}} \) is a dynamic weighting function with \( k =5\). \( \hat{x}_{0|t} = \frac{\hat{x}_t - \sqrt{1 - \bar{\alpha}_t} \cdot \varepsilon_{\theta}}{\sqrt{\bar{\alpha}_t}}\) is the intermediate denoising result through the reverse denoising at time step \(t\), where  \(\bar{\alpha}_t = \prod_{s=1}^{t} (1 - \beta_t)\) is the cumulative product coefficient of diffusion schedule. A wavelet transform \cite{burrus1998wavelets} is applied to both \(\hat{x}_{0|t}\) and \(x_0\),  yielding the low-frequency components \(\hat{x}_{0|t}^{l}, x_{0}^{l}\) and the high-frequency components \(\hat{x}_{0|t}^{h},x_{0}^{h}\). \( \mathcal{L}_{\text{low}} \) employs MSE \cite{wang2009mean} to enforce smoothness and maintain global consistency. In contrast, \( \mathcal{L}_{\text{high}} \) utilizes SSIM \cite{wang2004image} to calculate the high-frequency loss, enhancing the accurate recovery of local details and textures. 

\section{DriveWeather Dataset} 
Existing weather restoration datasets suffer from two critical limitations \cite{li2020all, guo2025onerestore}: (1) simplistic degradation embeddings that neglect real-world atmospheric effects such as weather-induced contrast reduction and background tonal shifts, and (2) constrained application scenarios focused on static objects (e.g., buildings, animals) rather than autonomous driving environments. To address these limitations, we propose DriveWeather dataset, a comprehensive driving scene dataset inspired by A-BDD \cite{assion2024bdd}. Our dataset encompasses seven challenging weather conditions critical to autonomous navigation: \textit{densefog, rainfog, overcast, rainstreaks, puddles, droplets, and glare}. Each condition features five severity levels (mild to extreme) to simulate progressive degradation intensity.

The dataset comprises 29,750 precisely aligned image pairs ($1280\times720 $ resolution), with 28,000 training samples and 1,750 test cases. Unlike A-BDD's original 63,700-image collection for segmentation, we implement stringent quality control: 1) Manual verification eliminates misaligned pairs caused by temporal inconsistencies in dynamic scenes. 2) Photometric calibration ensures accurate representation of weather-induced contrast variations. 3) Scenario diversity covers urban roads, highways, and intersections with moving vehicles, pedestrians, and traffic signs.

\textbf{Weather Data Types:} As shown in Figure \ref{figure:driverweather_dataset}, each weather condition includes five density levels (1-5).
\textit{\textbf{Densefog}} combines fog and overcast, using the scene's depth map to control the intensity of the opacity and blur effects.
\textit{\textbf{Overcast}} desaturates the image and replaces the sky with a gray tone using segmentation.
\textit{\textbf{Glare}} simulates light glare from the sky and ground shadows, with positions calculated from sun location and depth.
\textit{\textbf{Rainstreaks}} add rain streaks to the image using a particle system.
\textit{\textbf{Raindrops}} apply overcast and calculates reflections on surfaces using depth maps for intensity and position.
\textit{\textbf{Puddles}} simulate puddles with reflections. The puddle shapes are generated with Perlin noise and projected onto segmented streets.
\textit{\textbf{Rainfog}} combines overcast, rain streaks, wet streets, lens droplets, and fog.









\section{Experiments}
\subsection{Implementation Details.}
\label{training details}
We implement LCDiff using the PyTorch framework and adopt the same backbone architecture as previous approaches \cite{luo2023refusion}. During the diffusion process, we utilize DDIM sampling \cite{song2020denoising}. The model is trained on 4 NVIDIA A40 GPUs for 800K iterations, using the Adam optimizer with momentum parameters \( \beta_1 = 0.9 \) and \( \beta_2 = 0.999 \). The initial learning rate is set to \( 2 \times 10^{-4} \), and it is reduced according to a cosine decay schedule. An exponential moving average (EMA) strategy is applied for parameter updates, with a decay rate of 0.995. The noise schedule \( \beta_{t} \) ranges from 0.0001 to 0.02. For data augmentation, we use $256\times 256$ image patches, applying horizontal flipping and random rotations with fixed angles.

\subsection{Experimental Settings.}

In this section, we conduct extensive experiments on three all-weather restoration datasets: \textit{All-Weather} \cite{li2020all, valanarasu2022transweather}, Composite Degradation (CDD11) \cite{guo2025onerestore}, and our DriveWeather. 

To thoroughly evaluate the feasibility and effectiveness of the AWIR tasks, we compare our approach with two categories of state-of-the-art methods: general-purpose restoration models and all-in-one restoration methods. The general-purpose models include MPRNet \cite{zamir2021multi}, NAFNet \cite{chen2022simple}, Restormer \cite{zamir2022restormer}, and FSNet \cite{cui2023image}. All-in-one models include All-in-one \cite{li2020all}, TransWeather \cite{valanarasu2022transweather},  TKL\&MR \cite{chen2022learning}, WGWS-Net \cite{zhu2023learning}, $\text{WeatherDiff}_{64}$ \cite{ozdenizci2023restoring}, 
PromptIR \cite{potlapalli2024promptir}, AWRCP \cite{ye2023adverse}, DA-CLIP \cite{luo2024controlling}, MoFME \cite{zhang2024efficient}, OneRestore \cite{guo2025onerestore}, TransWeather-TUR \cite{wu2025debiased}, DPPD \cite{dppd} and MoCE-IR \cite{MoCEIR}. 
For a fair comparison under identical conditions, all competing methods in Table \ref{table:driveweather_result} were retrained on our DriveWeather dataset.

\begin{table*}[!t]
    \centering
    \renewcommand{\arraystretch}{1}
       \caption{Multiple weather restoration comparisons, including image desnowing, deraining, deraining \& dehazing. The 1st and 2nd best results are emphasized with \textbf{\color{reda}red} and \textbf{\color{myblue}blue}, respectively.}
       \vskip -0.1in 
    \setlength{\tabcolsep}{7pt}
    \renewcommand{\arraystretch}{1.2}
    \scalebox{0.96}{
    \begin{tabular}{p{0.25cm}|c|c|cc|cc|cc|cc|cc}
 \Xhline{1.2pt} 
        \multirow{2}{*}{\rotatebox[origin=c]{90}{Type}} & \multirow{2}{*}{Methods} 
          & \multirow{2}{*}{Venue} 
          & \multicolumn{2}{c|}{Snow100K-S} 
          & \multicolumn{2}{c|}{Snow100K-L}
          & \multicolumn{2}{c|}{RainDrop} 
          & \multicolumn{2}{c|}{Outdoor-Rain}
          & \multicolumn{2}{c}{\textbf{Average}} \\ 
        
          ~ & ~ & ~ & PSNR & SSIM & PSNR & SSIM & PSNR & SSIM & PSNR & SSIM & PSNR & SSIM  \\ \Xhline{0.9pt}

         \multirow{4}{*}{\rotatebox[origin=c]{90}{General}} & MPRNet \cite{zamir2021multi} & CVPR'21  & 34.97 & 0.945  & 29.76 & 0.894 & - & - & 28.03 & 0.919 & - & - \\

        ~ & NAFNet \cite{chen2022simple} & ECCV'22  & 34.79 & 0.949  & 30.06 & 0.901 & - & - & 29.59 & 0.902  & - & -\\

        ~ & Restormer \cite{zamir2022restormer} & CVPR'22  & 35.03 & 0.948  & 30.28 & 0.912 & 30.91 & 0.915 & 30.21 & 0.920 & 31.60 & 0.923 \\ 

        ~ & FSNet \cite{cui2023image} & PAMI'23  & 36.62 & 0.961  & 30.94 & 0.904 & 31.06 & \textbf{\color{myblue}0.934} & \textbf{\color{myblue}31.61} & 0.923 & 32.55 & 0.930 \\ \Xhline{0.9pt}  
        
         \multirow{13}{*}{\rotatebox[origin=c]{90}{All-in-one}} & All-in-one  \cite{li2020all} & CVPR'20 &- & -  & 28.33 & 0.882 & 31.12 &0.926 & 24.71 & 0.898  & - & -\\ 
         
          ~ & TransWeather \cite{valanarasu2022transweather} & CVPR'22 & 32.51 & 0.934 & 29.31 &  0.887 & 30.17 &0.915 & 28.83 & 0.900 & 30.20 & 0.909\\

         ~ & TKL\&MR \cite{chen2022learning} & CVPR'22 & 34.80 & 0.948 & 30.24 & 0.902 & 30.99 & 0.927 & 29.92 & 0.916 & 31.48 & 0.923\\
        
         ~ & WGWS-Net \cite{zhu2023learning} & CVPR'23 & 36.11 & 0.949 & 29.71 & 0.890 & 31.31 & 0.930 & 25.31 & 0.900 & 30.61 & 0.917\\

         ~ & $\text{WeatherDiff}_{64}$ \cite{ozdenizci2023restoring}  & PAMI'23 & 35.83 & 0.956 & 30.09 & 0.904 & 30.71 & 0.931 & 29.64 & 0.931& 31.56 & 0.930\\

        ~ & PromptIR \cite{potlapalli2024promptir} & NIPS'23  & 34.60 & 0.956  & 30.52 & 0.826 & 30.70 & 0.913 & 30.95 & 0.922  & 31.69 & 0.904\\ 
        
         ~ & AWRCP \cite{ye2023adverse} & ICCV'23 & \textbf{\color{myblue}36.92} &\textbf{\color{reda}0.965}  & \textbf{\color{myblue}31.92}  & \textbf{\color{myblue}0.934} & \textbf{\color{myblue}31.93} & 0.931 & 31.39& 0.932& \textbf{\color{myblue}33.04} & \textbf{\color{myblue}0.940}\\
         
        ~ & DA-CLIP \cite{luo2024controlling} & ICLR'24 & 32.65 & 0.869 & 30.02 &  0.762 & 29.03 &0.837 & 29.37& 0.870& 30.26 & 0.834 \\
        
         ~ & MoFME \cite{zhang2024efficient} & AAAI'24 & -  & - & 29.35 & 0.893 & 29.27 & 0.930 & 28.66  & \textbf{\color{myblue}0.940} & - & - \\
         
         ~ & OneRestore \cite{guo2025onerestore}  & ECCV'24 & 36.80 & 0.956 & 31.46 & 0.906 & 31.77 &0.931 & 30.49& 0.932 & 32.63 & 0.931\\

        ~ & TransWeather-TUR \cite{wu2025debiased} & AAAI'25 & 34.14 & 0.937 & 30.32 & 0.892 & 31.61 & 0.933 & 29.75 & 0.907 & 30.67 & 0.907 \\

        ~ & MoCE-IR \cite{MoCEIR} & CVPR'25 & 35.66 & \textbf{\color{myblue}0.957} & 30.49 & 0.913 & 30.92 & 0.917 & 30.04 & 0.906 & 31.78 & 0.923 \\ \cline{2-13}
        
          ~ &  \textbf{LCDiff}(Ours)  & - & \textbf{\color{reda} 37.61} & \textbf{\color{reda}0.965} & \textbf{\color{reda} 32.42} & \textbf{\color{reda} 0.935}
          &\textbf{\color{reda}32.62} & \textbf{\color{reda} 0.949} & \textbf{\color{reda} 32.13} & \textbf{\color{reda} 0.948}   & \textbf{\color{reda} 33.69}& \textbf{\color{reda} 0.949} \\  \Xhline{1.2pt} 

    \end{tabular}
  }
    \label{table:all_weather_result}
\end{table*}


\begin{figure*}[!t]
    \centering 
    \includegraphics[width=0.96\linewidth]{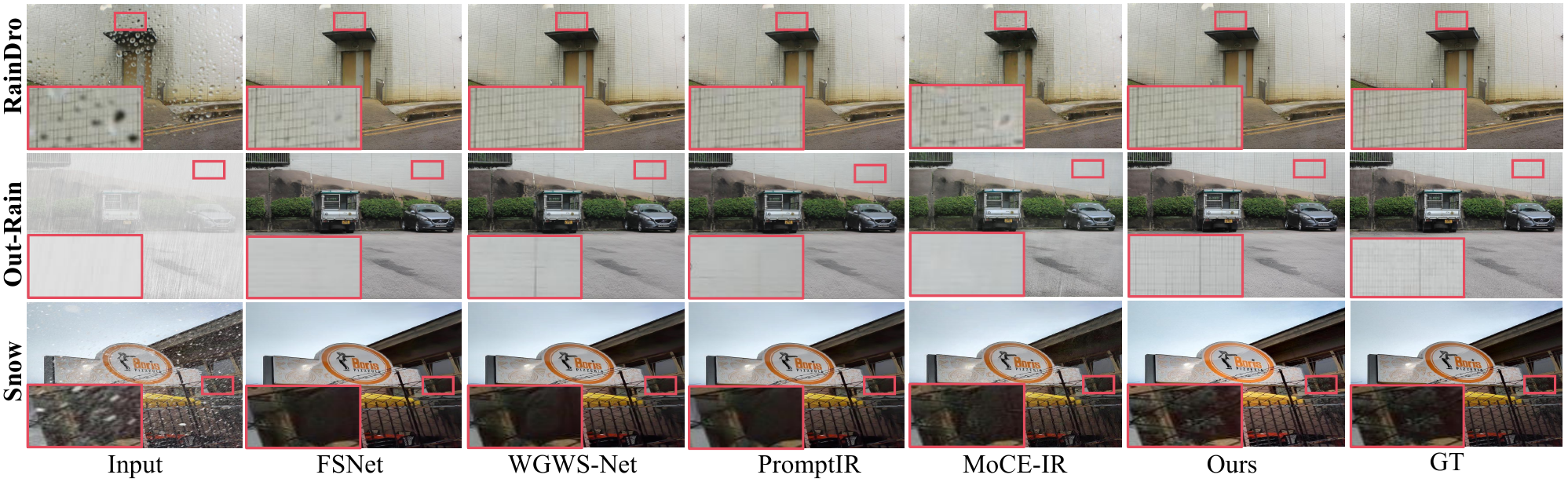}
    \caption{Visual results showcasing restoration performance on Snow100K \cite{liu2018desnownet}, Outdoor-Rain \cite{li2019heavy}, and RainDrop \cite{qian2018attentive} datasets.} 
    \label{figure:all_weather_result}
   
\end{figure*}

\begin{table*}[t!]
    \caption{Quantitative comparison of different methods on our proposed DriveWeather dataset.}
    \label{table:driveweather_result}
    \centering
    \setlength{\tabcolsep}{5pt}
    \renewcommand{\arraystretch}{1.2}
    \scalebox{0.9}{
    \begin{tabular}{c|cc|cc|cc|cc|cc|cc|cc|>{\columncolor{mygreen!10}}c >{\columncolor{mygreen!10}}c}
    \Xhline{1.2pt} 
    \multirow{2}{2cm}{\centering Methods \\ / Venue }  
        &  \multicolumn{2}{c|}{Restormer \cite{zamir2022restormer}} 
        & \multicolumn{2}{c|}{FSNet \cite{cui2023image}} 
        & \multicolumn{2}{c|}{PromptIR \cite{potlapalli2024promptir}} 
        & \multicolumn{2}{c|}{WGWS-Net \cite{zhu2023learning}} 
        & \multicolumn{2}{c|}{OneRestore \cite{guo2025onerestore}} 
        & \multicolumn{2}{c|}{DA-CLIP \cite{luo2024controlling}} 
        & \multicolumn{2}{c|}{MoCE-IR \cite{MoCEIR}} 
        &  \multicolumn{2}{c}{\textbf{LCDiff}}   \\  
        ~ 
        & \multicolumn{2}{c|}{CVPR'22} 
        & \multicolumn{2}{c|}{TPAMI'23} 
        & \multicolumn{2}{c|}{NeurIPS'23} 
        & \multicolumn{2}{c|}{CVPR'23} 
        & \multicolumn{2}{c|}{ECCV'24} 
        & \multicolumn{2}{c|}{ICLR'24} 
        & \multicolumn{2}{c|}{CVPR'25} 
        & \multicolumn{2}{c}{(Ours)}    \\  \hline
        
        Metrics  & PSNR & SSIM & PSNR & SSIM & PSNR & SSIM & PSNR & SSIM & PSNR & SSIM & PSNR & SSIM & PSNR & SSIM
        &  PSNR & SSIM    \\ \hline \hline 

        Densefog
        & 15.00 & 0.676
        & 22.45 &  0.901 
        & 22.43 & 0.900 
        & \textbf{\color{myblue}24.03} & 0.887 
        & 23.93 & 0.876
        & 23.32 & 0.886
        & 23.82 & \textbf{\color{myblue}0.913} 
        & \textbf{\color{reda}24.36} & \textbf{\color{reda}0.914} \\  

        Raindrop
        & 15.63 & 0.776
        & 30.84& \textbf{\color{myblue}0.961} 
        & 31.36& 0.956 
        & \textbf{\color{myblue}32.69} & 0.954 
        & 32.09 & 0.950
        & 32.45 & 0.936
        & 30.46 & 0.958 
        &\textbf{\color{reda}33.98}  & \textbf{\color{reda}0.967} \\

        Overcast 
        & 16.57 & 0.785
        & 33.03 & 0.967 
        & 36.02 & 0.980
        & 36.44 & 0.980
        & 36.55 & \textbf{\color{myblue}0.981}
        & 36.43 & 0.979
        & \textbf{\color{myblue}37.01} & 0.978 
        & \textbf{\color{reda}38.25}  & \textbf{\color{reda}0.984} \\  
        
        Puddles
        & 15.93 & 0.770
        & 33.50 & 0.960 
        & 33.67& 0.970 
        & 35.09 & 0.960 
        & 34.88 & 0.969
        & 34.89 & 0.964
        & \textbf{\color{myblue}35.25} & \textbf{\color{myblue}0.973} 
        & \textbf{\color{reda}35.90} & \textbf{\color{reda}0.977} \\  
        
        Rainfog
        & 14.76 & 0.676
        & 20.60 & 0.869 
        & 21.61 & 0.867 
        & 20.67 & \textbf{\color{myblue}0.881 }
        & 18.43 & 0.849
        & 19.60 & 0.868 
        & \textbf{\color{myblue}21.73} & 0.876
        & \textbf{\color{reda}22.45} & \textbf{\color{reda}0.884} \\

        Rainstreaks
        & 17.29 & 0.802
        & 26.12 & \textbf{\color{myblue}0.930} 
        & 27.61 & 0.925 
        & 27.44 & 0.920 
        & \textbf{\color{myblue}26.90} & 0.903
        & 26.84 & 0.871 
        & 25.15 & 0.921
        & \textbf{\color{reda}28.91} & \textbf{\color{reda}0.938} \\  
        
        Glare
        & 16.34 & 0.827
        & 24.39 & 0.956
        & 24.09 & 0.948 
        & 24.01 & 0.954 
        & 22.93 & 0.944
        & 22.15 & 0.946
        & \textbf{\color{reda}24.94} & \textbf{\color{reda}0.958}
        & \textbf{\color{myblue}24.90} & \textbf{\color{reda}0.958} \\ \hline \hline 
        
       \rowcolor{cyan!5} \textbf{Average} 
        & 15.93 & 0.758
        & 27.27& 0.934 
        & 28.11& 0.935
        & \textbf{\color{myblue}28.62} & 0.933
        & 27.96 & 0.938
        & 27.95 & 0.921 
        & 28.33 & \textbf{\color{myblue}0.939}
        & \textbf{\color{reda}29.82} & \textbf{\color{reda}0.946}
         \\  \Xhline{1.2pt} 
    \end{tabular}
}

\end{table*}


\subsection{Quantitative comparison.}

Our LCDiff tops the list on all benchmarks in Table~\ref{table:all_weather_result}, outperforming both general and all-in-one baselines. Unlike general models (e.g., FSNet~\cite{cui2023image}, Restormer~\cite{zamir2022restormer}) that perform poorly on mixed weather, LCDiff performs strongly on RainDrop and Outdoor-Rain, and outperforms AWRCP~\cite{ye2023adverse} by +1.97\% and +0.96\% in PSNR and SSIM, respectively. Prompt-driven methods (OneRestore~\cite{guo2025onerestore}, PromptIR~\cite{potlapalli2024promptir}) benefit from the VL prior but are inconsistent with overlap degradation. In the DriveWeather dataset (Table~\ref{table:driveweather_result}), LCDiff performs best across 6 of 7 weather types, achieving +5.3\% and +0.7\% higher peak signal-to-noise ratio (PSNR) and SSIM than MoCE-IR~\cite{MoCEIR}, respectively. LCDiff's improvements are particularly significant on the rain and fog (+0.72 dB) and rain streak (+1.3 dB) datasets. On the glare dataset (Glare), LCDiff slightly outperforms in SSIM but lags behind by 0.04 dB in PSNR, indicating room for improvement in highlight suppression. On the CDD11 dataset (Table~\ref{table:cdd}), LCDiff achieves the best overall average, with PSNR and SSIM higher than MoCE-IR by +1.6 dB and +0.6 dB, respectively, and shows significant improvements on both the multi-degraded (L+H+R, up to +1.1 dB) and single-shot low-light (L, +1.55 dB) datasets.

\subsection{Visual Comparison.}
\begin{figure*}[t]
    \centering 
    \includegraphics[width=0.96\linewidth]{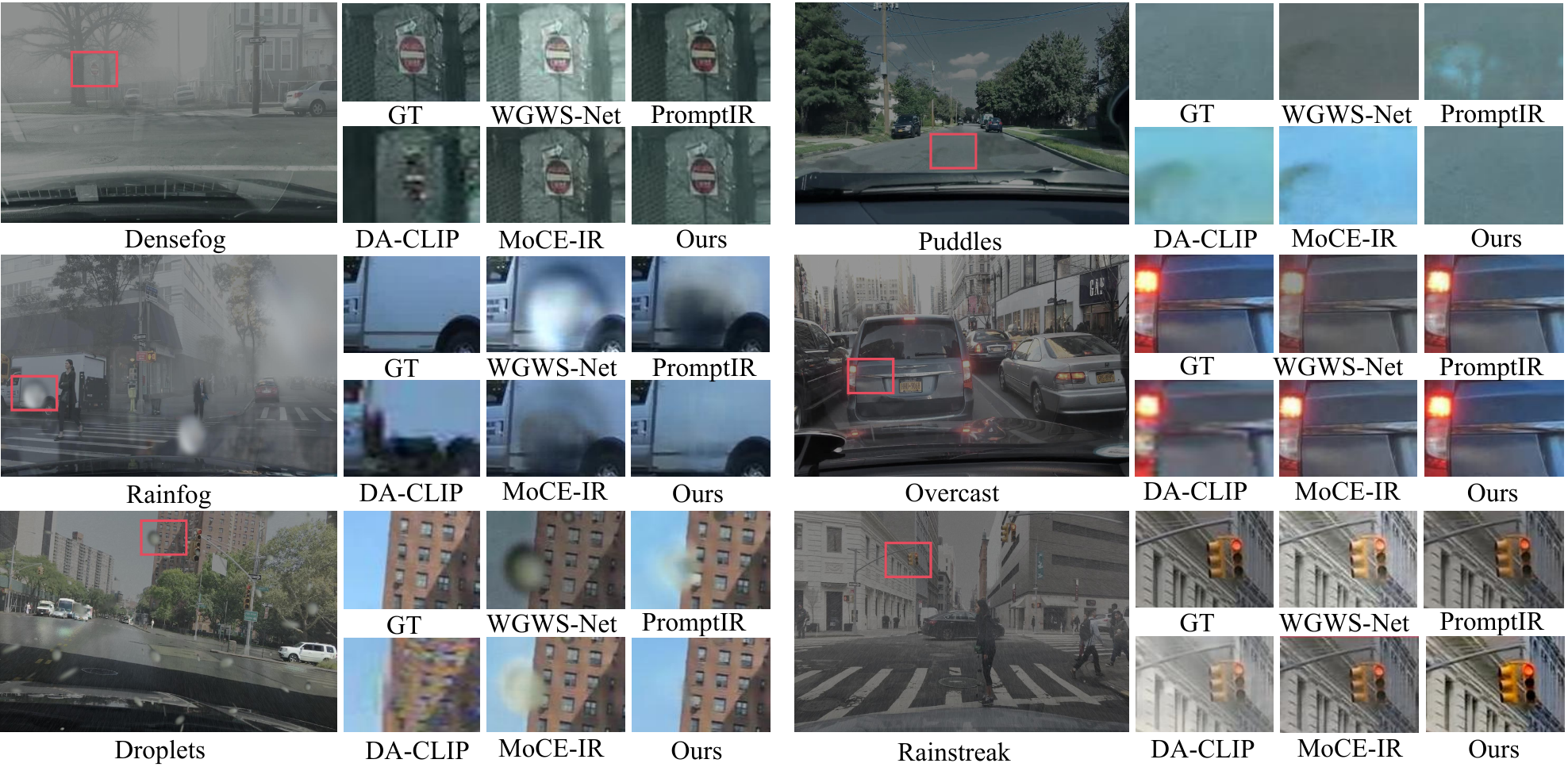}
    \caption{Visual demonstration of the restoration performance on our DriveWeather dataset.} 
    \label{figure:driveweather_result}
\end{figure*}

\begin{figure*}[t]
    \centering 
    \includegraphics[width=0.96\linewidth]{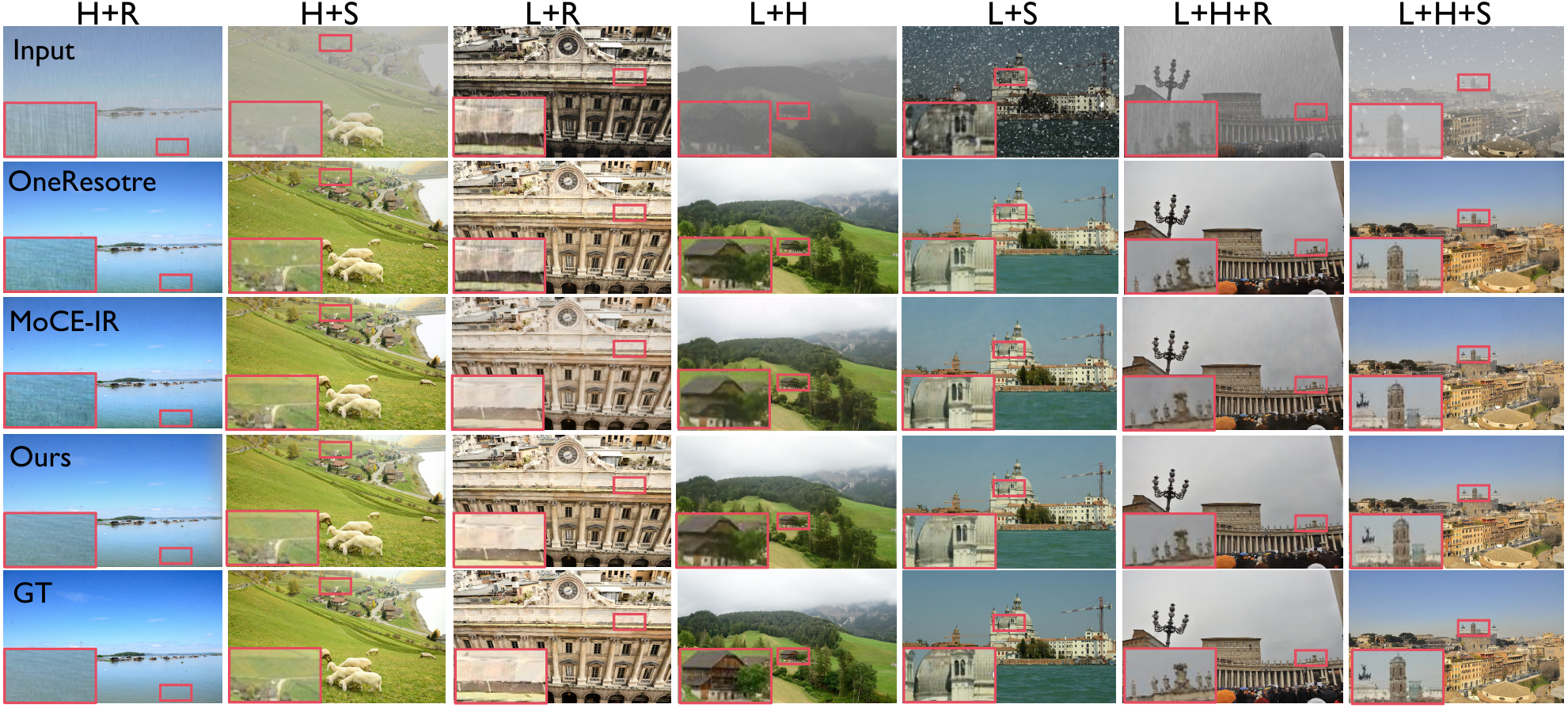}

    \caption{Visual demonstration of the restoration performance on CDD11 dataset.} 
    \label{figure:cdd11_result}
   
\end{figure*}

\begin{figure*}[!t]
    \centering 
    \includegraphics[width=0.96\linewidth]{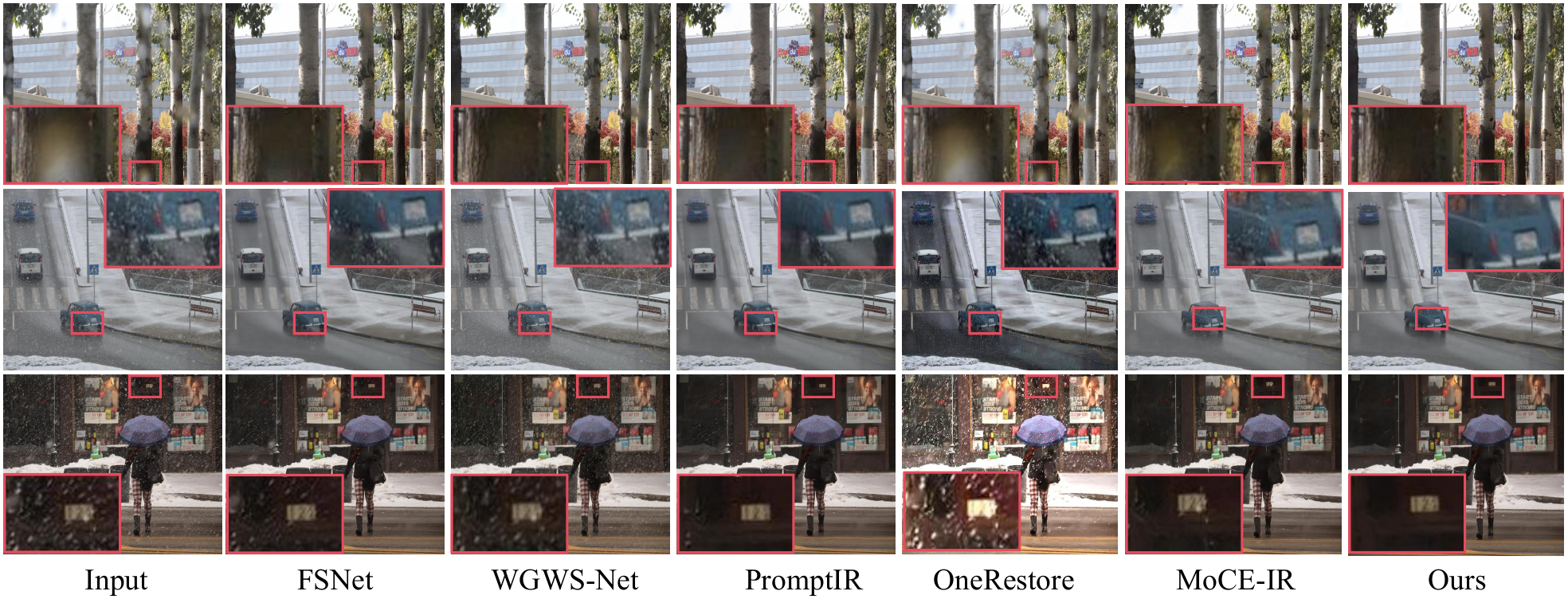}
    \caption{Visual comparison of real-world adverse weather samples, including real rain images from the RainDS \cite{quan2021removing} dataset and real snow images from the Snow100K \cite{liu2018desnownet} dataset.} 
    \label{figure:real_results}
\end{figure*}

As shown in Figs~\ref{figure:all_weather_result},~\ref{figure:driveweather_result}, and~\ref{figure:cdd11_result}, our LCDiff consistently produces sharper structures and more natural textures across diverse degradation types. For instance, Figure \ref{figure:all_weather_result} demonstrates that LCDiff removes artifacts better than FSNet, WGWS-Net, and MoCE-IR while preserving details and illumination. Figure~\ref{figure:driveweather_result}  highlights LCDiff’s ability to restore both global and local details on DriveWeather, maintaining natural lighting. In Figure \ref{figure:cdd11_result}, LCDiff achieves clearer structures and finer textures, especially in challenging cases (L+H+R, L+H+S). Compared to OneRestore and MoCE-IR, our results show fewer artifacts, better color fidelity (e.g., cleaner skies, sharper edges, and more vivid grass). Finally, we present a visual comparison of image restoration in real-world scenarios (Figure~\ref{figure:real_results}). In rainy and snowy conditions, while methods like FSNet \cite{cui2023image}, WGWS-Net \cite{zhu2023learning}, and PromptIR \cite{potlapalli2024promptir} remove raindrops, they struggle with fine texture preservation and consistent lighting. In contrast, our method not only removes raindrops more thoroughly but also restores subtle scene details.

\begin{table*}[!t]
    \centering
    \renewcommand{\arraystretch}{1.2}
       \caption{Comparison to state-of-the-art on composited degradations in the CDD11 dataset.}
      
    \setlength{\tabcolsep}{2pt}
    \renewcommand{\arraystretch}{1.2}
    \scalebox{0.9}{
    \begin{tabular}{c|c|cc|cc|cc|cc|cc|cc|cc|cc|cc|cc|cc|cc}
 \Xhline{1.2pt} 
         \multirow{2}{*}{Methods} 
          & \multirow{2}{*}{Venue} 
          & \multicolumn{8}{c|}{\textit{CDD11-Single}} 
          & \multicolumn{10}{c|}{\textit{CDD11-Double}}
          & \multicolumn{4}{c|}{\textit{CDD11-Triple}} 
          & \multicolumn{2}{c}{\multirow{2}{*}{Avg}} \\ 
          
       \cline{3-24}
         
           ~ & ~ & \multicolumn{2}{c}{Low(L)}   & \multicolumn{2}{c}{Haze(H)}  &\multicolumn{2}{c}{Rain(R)}  & \multicolumn{2}{c|}{Snow(S)}  &\multicolumn{2}{c}{L+H}  & \multicolumn{2}{c}{L+R}  & \multicolumn{2}{c}{L+S} & \multicolumn{2}{c}{H+R}  & \multicolumn{2}{c|}{H+S}  & \multicolumn{2}{c}{L+H+R}  &  \multicolumn{2}{c|}{L+H+S} &  \\ 
           \hline \hline 
          AirNet  &  CVPR'22 & 
         24.83 &\cc{.778} & 
         24.21 &\cc{.951} & 
         26.55&\cc{.891} & 
         26.79&\cc{.919}
        & 23.23&\cc{.779} & 
        22.82&\cc{.710} & 
        23.29&\cc{.723} & 
        22.21&\cc{.868} & 
        23.29&\cc{.901}
        & 21.80&\cc{.708} & 
        22.24&\cc{.725} & 
        23.75&\cc{.814}\\

         PromptIR  & NIPS'23 & 
        26.32&\cc{.805} & 
        26.10&\cc{.969} & 
        31.56&\cc{.946} & 
        31.53&\cc{.960} & 
        24.49 &\cc{.789} & 
        25.05&\cc{.771} & 
        24.51&\cc{.761} & 
        24.54&\cc{.924} & 
        23.70&\cc{.925} & 
        23.74&\cc{.752} & 
        23.33&\cc{.747} & 
        25.90&\cc{.850}\\

         WGWS-Net   & CVPR'23 & 
        24.39&\cc{.774} & 
        27.90&\cc{.982} & 
        33.15&\cc{.964} & 
        34.43&\cc{.973} & 
        24.27&\cc{.800} & 
        25.06&\cc{.772} & 
        24.60&\cc{.765} & 
        27.23&\cc{.955} & 
        27.65&\cc{.960} & 
        23.90&\cc{.772} & 
        23.97&\cc{.771} & 
        26.96&\cc{.863} \\ 

         WeatherDiff  & PAMI'23 & 
        23.58&\cc{.763} & 
        21.99&\cc{.904} & 
        24.85&\cc{.885} & 
        24.80&\cc{.888} & 
        21.83&\cc{.756} & 
        22.69&\cc{.730} & 
        22.12&\cc{.707} & 
        21.25&\cc{.868} & 
        21.99&\cc{.868} & 
        21.23&\cc{.716} & 
        21.04&\cc{.698} & 
        22.49&\cc{.799} \\ 

        OneRestore  & ECCV'24 & 
       26.48 &\cc{.826} & 
       32.52&\cc{.990} & 
       33.40&\cc{.964 }& 
       34.31&\cc{.973} & 
       25.79&\cc{\textbf{\color{reda}.822}} &
       25.58&\cc{\textbf{\color{myblue}.799}} & 
       25.19&\cc{.789} & 
       \textbf{\color{myblue}29.99}&\cc{.957} & 
       \textbf{\color{myblue}30.21}&\cc{.964} & 
       24.78&\cc{.788} & 
       24.90&\cc{\textbf{\color{myblue}.791}} & 
       28.47&\cc{.878} \\

        DPPD & TIP’25 & 
        27.22 & \cc{.820} &
        31.08 & \cc{.988} &
        \textbf{\color{myblue}34.53} & \cc{.968} &
        35.48 & \cc{.978} & 
        \textbf{\color{reda}26.29} & \cc{.812} &
        \textbf{\color{reda}26.36} & \cc{.796} & 
        \textbf{\color{myblue}26.11} & \cc{.789} &
        29.87 & \cc{.958} & 
        29.64 & \cc{.963} & 
        25.30 & \cc{.782} & 
        25.37 & \cc{.781} &
        28.84 & \cc{.875} \\
        
        MoCE-IR & CVPR'25 & 
        \textbf{\color{myblue}27.26} & \cc{\textbf{\color{myblue}.824}} & 
        \textbf{\color{myblue}32.66} & \cc{\textbf{\color{myblue}.990}} & 
        34.31& \cc{\textbf{\color{myblue}.970}} & 
        \textbf{\color{myblue}35.91} & \cc{\textbf{\color{myblue}.980}} & 
        26.24& \cc{.817} &
        \textbf{\color{myblue}26.25} & \cc{\textbf{\color{reda}.800}} &
        26.04 & \cc{\textbf{\color{myblue}.793}} & 
        29.93&\cc{\textbf{\color{myblue}.964}} & 
        30.19 &\cc{\textbf{\color{myblue}.970}}& 
        \textbf{\color{myblue}25.41}&\cc{\textbf{\color{myblue}.789}} &
        \textbf{\color{myblue}25.39}&\cc{.790} & 
        \textbf{\color{myblue}29.05}&\cc{\textbf{\color{myblue}.881}}
            \\
            
        \hline  \hline 
        

       \textbf{LCDiff} (Ours)  & - & 
       \textbf{\color{reda}28.81} & \cc{\textbf{\color{reda}.849}} & 
       
      \textbf{\color{reda}33.32} & \cc{\textbf{\color{reda}.992}} & 
      
      \textbf{\color{reda}35.17} & \cc{\textbf{\color{reda}.970}} & 
      
      \textbf{\color{reda}35.95} & \cc{\textbf{\color{reda}.978}} & 
      
      \textbf{\color{myblue}26.28}& \cc{\textbf{\color{myblue}.818}} & 
      
       26.24 &  \cc{.797} &  
      \textbf{\color{reda}26.49} & \cc{\textbf{\color{reda}.810}} & 
      
      \textbf{\color{reda}31.05} & \cc{\textbf{\color{reda}.967}} & 
      
      \textbf{\color{reda}30.27} & \cc{\textbf{\color{reda}.971}} &
      
      \textbf{\color{reda}25.48} & \cc{\textbf{\color{reda}.790}} &
      
      \textbf{\color{reda}25.69} &  \cc{\textbf{\color{reda}.815}} & 
      
      \textbf{\color{reda}29.52} & \cc{\textbf{\color{reda}.887}}\\  \Xhline{1.2pt} 
    \end{tabular}
  } 
    \label{table:cdd}
\end{table*}

\begin{table}[t]
    \centering
    \caption{Ablation studies on the various components.}
  
    \label{table_ablation_com}
    \setlength{\tabcolsep}{4pt}
    \renewcommand{\arraystretch}{1.2}
    \scalebox{1}{
    \begin{tabular}{l|cc|cc|cc}
    \Xhline{1.2pt} 
        \rowcolor{mygray} &  \multicolumn{2}{c|}{ \textbf{Snow (Avg)} }
        &  \multicolumn{2}{c|}{ \textbf{Outdoor-Rain} }  & \multicolumn{2}{c}{ \textbf{RainDrop} } \\ \cline{2-7}
        \rowcolor{mygray}
        \multirow{-2}*{Method}   & PSNR  & SSIM & PSNR  & SSIM  & PSNR & SSIM  \\ \hline \hline

          Only LCDN &  31.42  & 0.906 &  29.80  & 0.915  & 30.01 & 0.919 \\
          
          LCDN w/o FCRM & 31.10 & 0.902 & 29.54 & 0.908 & 29.66 & 0.917 \\
            
        LCDN w/o LRM  & 27.20 & 0.852 & 25.64 & 0.832 & 26.57 & 0.836 \\
          
          Only LGDM &  32.20  & 0.923 &  29.96  & 0.926  & 31.09 & 0.920 \\ \hline

          LCDN+LGDM &  34.66  & 0.942 &  31.74   & 0.940  & 31.99 & 0.941 \\
          
          w/o \(\mathcal{L}_{\text{dts}}\) &  34.39  & 0.938 & 31.62 & 0.937 & 31.84 & 0.936 \\
          
          w/ condition &  34.97  & 0.947 & 31.82 & 0.946 & 32.23 & 0.946 \\
          
        \rowcolor{mygray}
        \scalebox{1.5}{{$\star$}} LCDiff &  \textbf{35.01} & \textbf{0.950} & \textbf{32.13} & \textbf{0.948} & \textbf{32.62} & \textbf{0.949} \\
          
          \Xhline{1.2pt} 
        \end{tabular}
        }

\end{table}

\begin{figure}[!t]
    \centering 
    \includegraphics[width=0.96\linewidth]{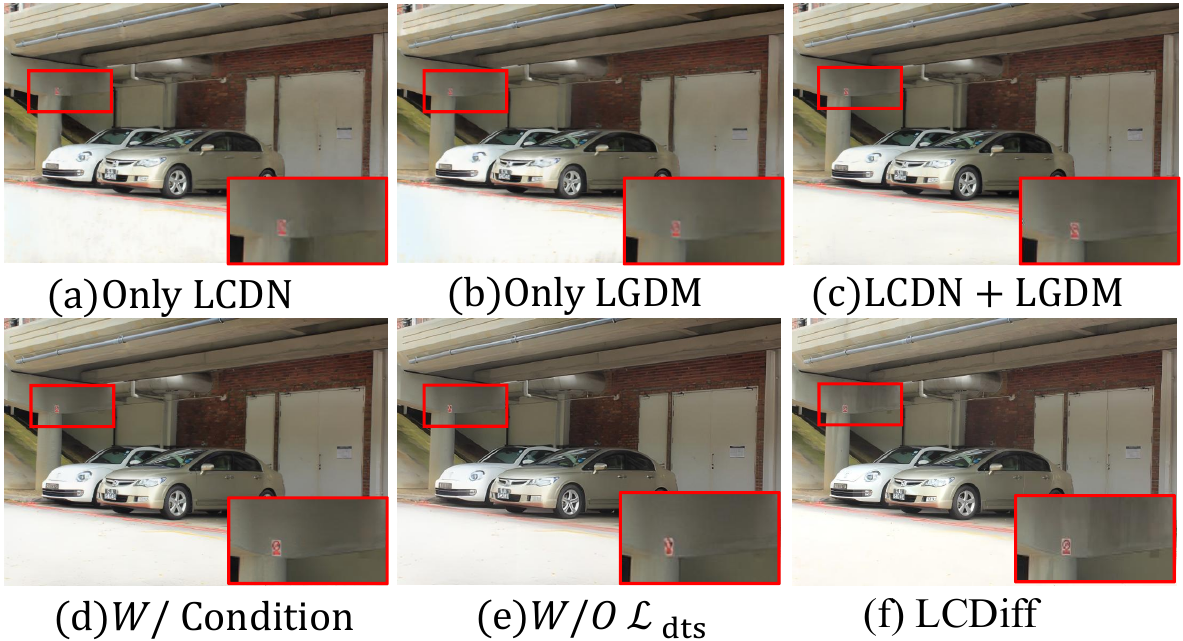}
      
        \caption{Visual ablation studies for our LCDiff.} 
    \label{figure:ablation_lcdn}
   
\end{figure}

\begin{figure}[!t]
    \centering 
    \includegraphics[width=0.96\linewidth]{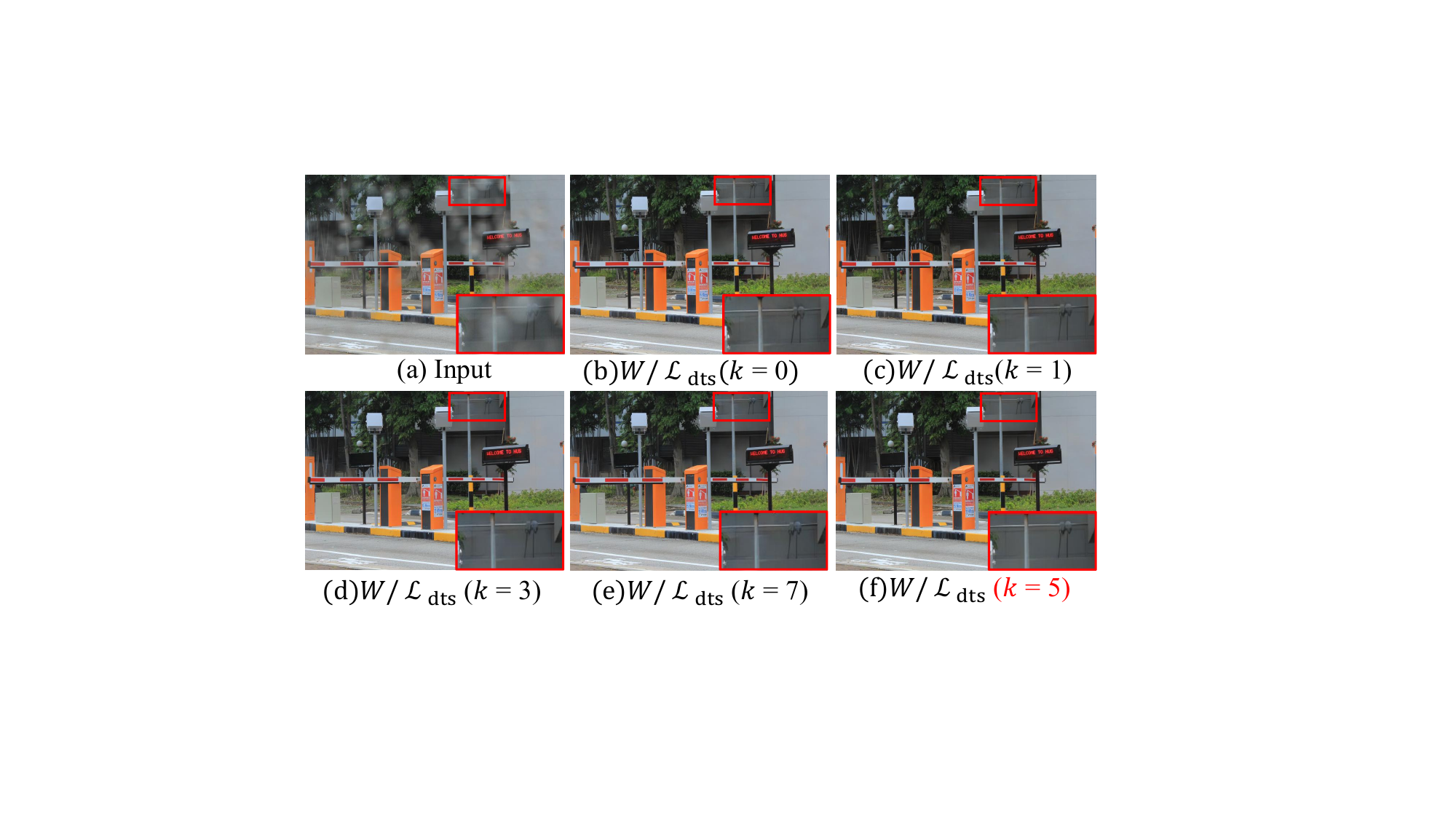}
    \caption{Visual results of $k$ values for \(\mathcal{L}_{\text{dts}}\).} 
    \label{figure:ablation_loss}

\end{figure}

\section{Ablation Study}
In this subsection, we perform ablation studies to evaluate the contributions of individual components in LCDiff on the all-weather dataset.

\begin{figure}[!t]
    \centering 
    \includegraphics[width=0.98\linewidth]{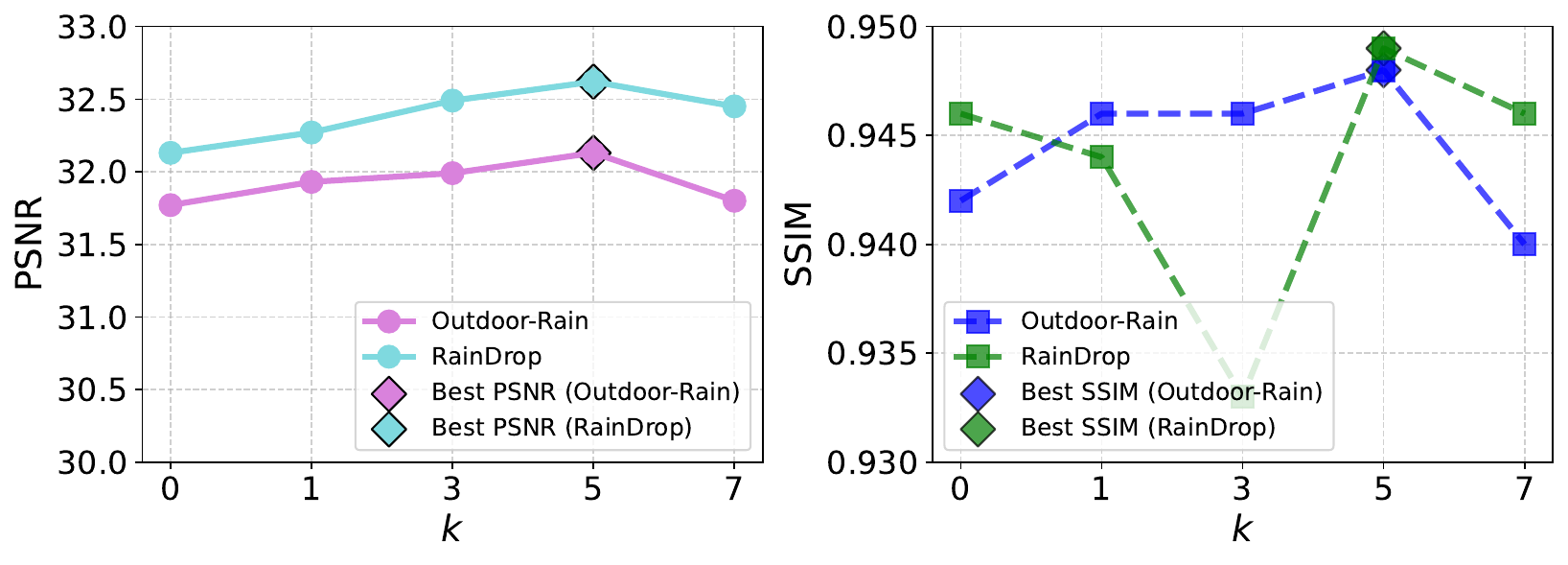}
    \caption{Ablation experiments with $k=0,1,3,5,7$.} 
    \label{figure:lossfunction}
   
\end{figure}

\textbf{Effectiveness of Each Components.}
As summarized in Tab.~\ref{table_ablation_com}, each module contributes distinctly to the final performance. 
\textbf{(i Decomposition matters.} Using only LCDN already yields strong results, while ablating its LRM (LCDN w/o LRM) causes the largest degradation across datasets, confirming that removing luminance–domain degradation is the primary driver of structural recovery. In contrast, removing FCRM (LCDN w/o FCRM) leads to a milder drop, indicating FCRM’s role as a color–consistency regulator that stabilizes chroma and suppresses bleed/oversmoothing. 
\textbf{(ii) Generative refinement helps but is not enough alone.} LGDM-only improves over LCDN on texture–heavy cases (e.g., Snow, RainDrop) by hallucinating high-frequency details, yet it is less stable on complex degradations when used without a deterministic front-end. Combining LCDN+LGDM unlocks clear gains on all subsets by coupling deterministic decontamination with generative detail synthesis. 
\textbf{(iii) Loss scheduling and conditioning are critical.} Removing the \(\mathcal{L}_{\text{dts}}\) term reduces both PSNR and SSIM, evidencing that the dynamic low\(\rightarrow\)high frequency curriculum is essential for balanced restoration. Adding conditional guidance (luminance cues) further improves fidelity and coherence by constraining the reverse process to a physically meaningful manifold. 
Overall, the LCDiff achieves the best scores consistently, delivering sharper textures, cleaner structures, and more stable colors across Snow, Outdoor-Rain, and RainDrop. Figure~\ref{figure:ablation_lcdn} provides a visual comparison of the restoration effects across different configurations.

\textbf{Effect of \(k\) value for \(\mathcal{L}_{\text{dts}}\).}
We conducted a series of experiments to investigate the impact of the hyperparameter \(k\) in \(\mathcal{L}_{\text{dts}}\) on model performance. As shown in Figure~\ref{figure:lossfunction}, varying \(k\!\in\!\{0,1,3,5,7\}\) reveals a clear trend on Outdoor-Rain and RainDrop: PSNR steadily increases up to \(k\!=\!5\) and then drops at \(k\!=\!7\); SSIM attains its best (or near-best) value likewise at \(k\!=\!5\). This behavior follows the weighting schedule \(\omega_t = e^{-k t/T}\): small \(k\) (\(0\!-\!1\)) keeps \(\omega_t\!\approx\!1\), over-emphasizing low-frequency MSE and causing over-smoothing; moderate \(k\) (\(3\!-\!5\)) smoothly shifts supervision toward high-frequency SSIM, stabilizing global structure early and refining textures later; excessively large \(k=7\) over-weights late-stage high-frequency terms, leading to over-sharpening/ringing and a metric decline. We therefore adopt \(k\!=\!5\) by default, which aligns with DDPM’s frequency dynamics (high frequencies vanish earlier than low) and yields the best balance between structural fidelity and fine-detail recovery. Figure~\ref{figure:ablation_loss} confirms that incorporating the loss \(\mathcal{L}_{\text{dts}}\) into the diffusion model enhances both the quality and consistency of restored images.

\begin{figure}[!t]
    \centering 
    \includegraphics[width=0.98\linewidth]{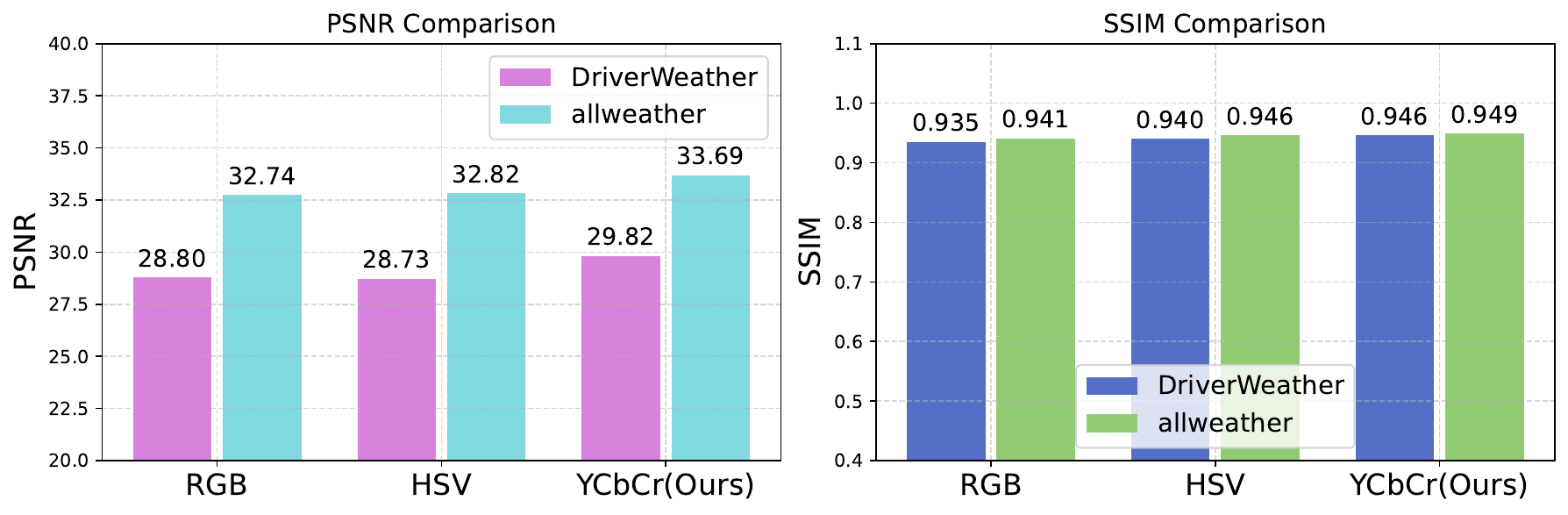}
    \caption{Performance of our method under different color-space decompositions (RGB, HSV, YCbCr) on two benchmarks (DriveWeather and All-Weather). YCbCr consistently delivers the highest PSNR/SSIM, validating that decoupling luminance and chrominance aligns with weather-degradation statistics.} 
    \label{figure:diff_color_space}
\end{figure}

\textbf{Effect of Color Space on Restoration.}
As shown in Figure~\ref{figure:diff_color_space}, decomposing in \textbf{YCbCr} yields the best results on both benchmarks. 
On DriveWeather, YCbCr delivers approximately +3.5\%/+3.8\% PSNR gains over RGB/HSV, and about +1.2\%/+0.6\% SSIM gains. 
On All-Weather, YCbCr yields roughly +2.9\%/+2.7\% PSNR and +0.9\%/+0.3\% SSIM improvements compared to RGB/HSV. \textbf{RGB.} Intensity and chroma are entangled across \(R/G/B\); haze/rain/glare corrupt all channels, causing denoising–color-correction gradient conflict and enlarging the diffusion search space. \textbf{HSV.} The \((H,S,V)\) mapping is nonlinear and discontinuous (hue wrap; ill-defined at low \(S\)), so small RGB perturbations induce large, non-Gaussian errors in \(H/S\), yielding unstable chroma despite a separate \(V\). \textbf{YCbCr (Ours).} A linear luma–chroma split concentrates degradation in \(Y\) while \(Cb/Cr\) remain low-variance. This matches illumination physics, preserves Gaussianity under the linear transform, and—by conditioning diffusion on \(Y\)—shrinks the denoising space to improve texture recovery and convergence, hence higher PSNR/SSIM.

\begin{table}[!t]
    \centering
       \caption{Comparison of different diffusion configurations on multiple weather conditions.}
    \setlength{\tabcolsep}{1.1pt}
    \renewcommand{\arraystretch}{1.4}
    \scalebox{0.96}{
    \begin{tabular}{l|cc|cc|cc|cc|c}
 \Xhline{1.2pt} 
           ~   & \multicolumn{2}{c}{Snow-S}  &\multicolumn{2}{c}{Snow-L}  & \multicolumn{2}{c}{RainDrop}  &\multicolumn{2}{c}{OutdoorRain} & Time  \\ \hline 

        Y-Diff  & 36.97 & \cc{.956} & 31.79 &\cc{.930} & 31.97 &\cc{.936} & 31.88&\cc{.924} & 0.46s \\ 
        
        Y/CBCR-Diff & 37.48 &\cc{.961} & 32.33 &\cc{.927} & 32.39 &\cc{.936} & 32.08 &\cc{.931} & 1.16s \\ \hline

        LCDiff(Ours) & \textcolor{red}{37.61} &\cc{\textcolor{red}{.965}} & \textcolor{red}{32.42} &\cc{\textcolor{red}{.935}} & \textcolor{red}{32.62} &\cc{\textcolor{red}{.949}} & \textcolor{red}{32.13} &\cc{\textcolor{red}{.948}} & \textcolor{red}{0.61s} \\
         
 \Xhline{1.2pt} 
    \end{tabular}
  }
    \label{table:doublediffusion}
\end{table}

\begin{figure}[!t]
    \centering 
    \includegraphics[width=0.98\linewidth]{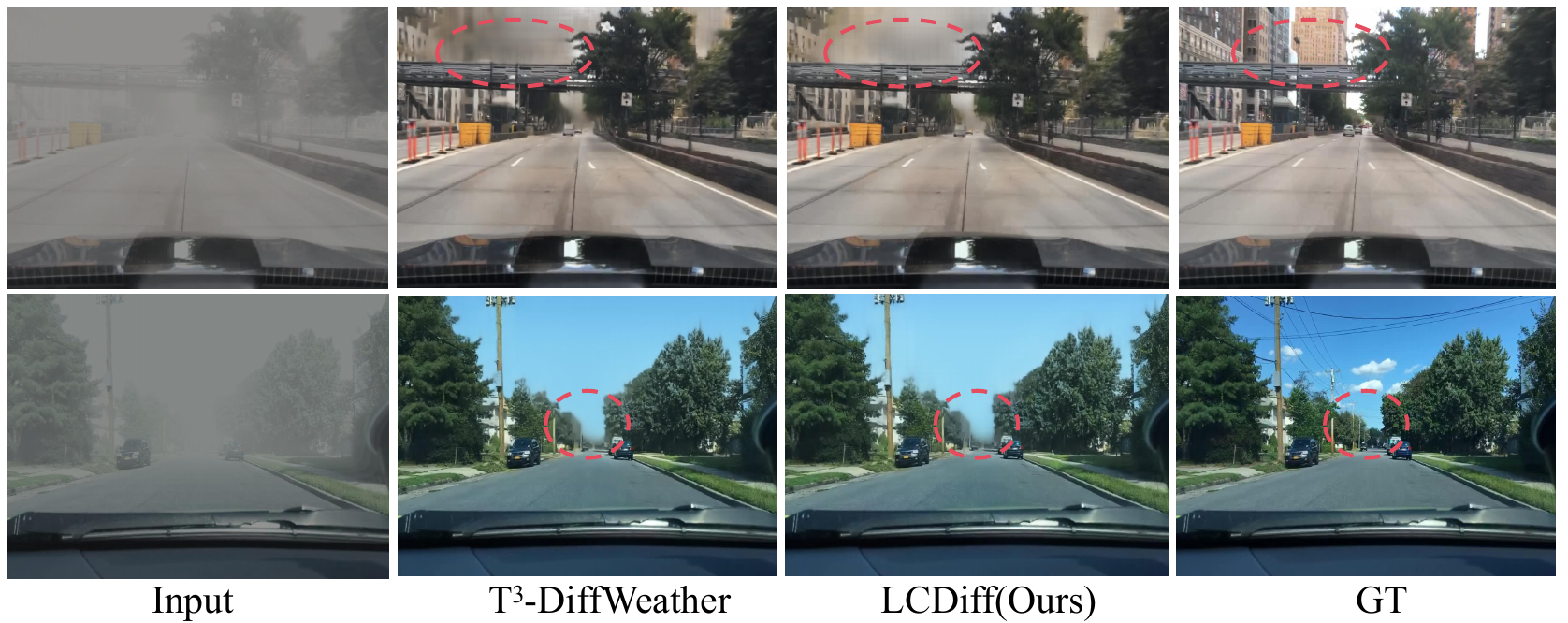}
    \caption{Visual results of the failure cases.} 
    \label{figure:failcases}
\end{figure}

\textbf{Ablation on Double-Diffusion Strategy}
To investigate whether applying separate diffusion models to luminance and chrominance channels would yield performance gains, we conduct an ablation comparing three configurations: (1) \textbf{Y-Diff}, which performs diffusion only on the Y channel; (2) \textbf{Y/CBCR-Diff}, which applies separate diffusion models to both Y and Cb/Cr channels; and (3) our proposed unified model \textbf{LCDiff}. As shown in Table~\ref{table:doublediffusion}, Y/CBCR-Diff brings marginal improvements over Y-Diff (e.g., +0.51 dB PSNR on Snow-S), but at the cost of \textbf{2.5× slower inference time} (1.16s vs. 0.46s). This clearly indicates diminishing returns from full chrominance diffusion. Chrominance channels (Cb/Cr) are relatively stable under most weather conditions, contributing less to structural degradation. Performing full diffusion on them introduces unnecessary complexity and risk of \textbf{color inconsistency}, as evident from the non-monotonic SSIM results in Y/CBCR-Diff (e.g., SSIM drops on Snow-L from .930 to .927 compared to Y-Diff). Our LCDiffusion model strikes a better balance by employing selective conditioning on the chrominance components, instead of separate full diffusion.

\subsection{Limitation and Failure Cases}
Despite its impressive performance, LCDiff shares a limitation common to current state-of-the-art models: as shown in Figure \ref{figure:failcases}, in heavily foggy scenes, the distant view is often obscured by dense fog, leading to significant information loss in the degraded image. This limitation hinders the restoration of fine details in distant backgrounds, which remains an open challenge for future research.

\section{Conclusion}
We introduced LCDiff, a new method designed to address image degradation under unknown weather conditions. By decomposing images into luminance and chrominance components in the YCbCr color space, LCDiff effectively removes degradation while preserving color consistency. Additionally, our luminance-guided diffusion model enhances restoration, enabling adaptive refinement of both low- and high-frequency details. Extensive experiments show that LCDiff outperforms the related methods across multiple datasets, delivering superior performance in diverse weather scenarios.

\ifCLASSOPTIONcaptionsoff
  \newpage
\fi

\bibliographystyle{IEEEtran}
\bibliography{main}

\vspace{-20pt}

\end{document}